%% file: main.tex
\let\TeXyear\year
\documentclass{IEEEaccess}

\let\year\TeXyear

\usepackage{amsmath,amssymb,amsfonts}
\usepackage{algorithmic}
\usepackage{graphicx}
\usepackage{textcomp}
\usepackage[numbers]{natbib}
\usepackage{marginnote}
\usepackage{times,latexsym}
\usepackage{url}
\usepackage{amsthm}
\usepackage[utf8]{inputenc}
\usepackage[normalem]{ulem}
\usepackage[english]{babel}
\usepackage[style=american]{csquotes}
\usepackage{subcaption}
\usepackage{tabularx}
\usepackage{multirow}
\usepackage{xcolor}
\usepackage{dirtytalk}
\usepackage{hyperref}
\usepackage{booktabs}
\usepackage{mathtools}
\usepackage{siunitx}
\usepackage[normalem]{ulem}

\usepackage{xspace,mfirstuc,tabulary}

\input{src/commands.tex}

\newcommand\hl[1]{#1}

\begin{document}
\history{}
\doi{}
\title{Assessing the impact of contextual information in hate speech detection}
\input{src/authors.tex}
\markboth
{Author \headeretal: Preparation of Papers for IEEE TRANSACTIONS and JOURNALS}
{Author \headeretal: Preparation of Papers for IEEE TRANSACTIONS and JOURNALS}

\corresp{Corresponding author: Juan Manuel Pérez (e-mail: jmperez at dc.uba.ar).}

\begin{abstract}
  \input{src/00_abstract.tex}
\end{abstract}


\titlepgskip=-15pt
\maketitle

\section{Introduction}

\input{src/01_intro.tex}

%
%
\section{Previous Work}
\input{src/02_previous.tex}

\section{Definition of Hate Speech}
\input{src/03_hate_speech_definition.tex}
\section{Corpus}
\input{src/04_corpus.tex}

\section{Classification Experiments}
\input{src/05_classifiers.tex}
\section{Results}
\input{src/06_results.tex}

\section{Discussion}
\input{src/07_discussion.tex}

\section{Conclusions}
\input{src/08_conclusions.tex}


\section*{Availability of Data and Material}
We make our corpus available at the huggingface hub \footnote{\url{https://huggingface.co/datasets/piubamas/contextualized_hate_speech}}. For the sake of reproducibility and also for further research, we will release the anonymized annotations (as suggested by \citet{basile2020s}) in addition to the aggregated dataset. The annotation guidelines will be publicly available upon publication of this paper.
%
%

\section*{Acknowledgments}

The authors would like to thank the annotators who worked to ensure the accuracy and quality of the data used in this study. Their dedication and hard work were instrumental in the success of this project. Thanks to A. Silva, G. Clerici, G. Damill, D. Valado, F. de Sanctis, L. Prats.

We would also like to thank Dr. Eugenia Mitchelstein, who provided valuable insights and suggestions that helped shape the direction of this research.

This research work was supported by grants for interdisciplinary research projects evaluated and funded by the Universidad de Buenos Aires (PIUBAMAS-2020-3 and PIUBA-2022-04-02). We would also like to thank CONICET and Universidad Torcuato Di Tella for their support.

\bibliography{biblio}
\bibliographystyle{IEEEtranN}

\include{src/biographies.tex}
\section{Supplemental material}
\input{src/appendix.tex}

\EOD
\end{document}

%% file: src/commands.tex
\newcolumntype{P}[1]{>{\centering\arraybackslash}p{#1}}
\newcommand{\mbf}[1]{\mathbf{#1}}

\newcommand{\mc}[2]{\multicolumn{#1}{c}{#2}}

\newcommand{\beto}[0]{BETO}

%% file: src/authors.tex
\author{
    \uppercase{Juan Manuel Pérez}\authorrefmark{1},
    \uppercase{Franco Luque}\authorrefmark{2},
    \uppercase{Demian Zayat}\authorrefmark{7},
    \uppercase{Martín Kondratzky}\authorrefmark{10},
    \uppercase{Agustín Moro}\authorrefmark{6, 8},
    \uppercase{Pablo Santiago Serrati}\authorrefmark{6, 9},
    \uppercase{Joaquín Zajac}\authorrefmark{6, 11},
    \uppercase{Paula Miguel}\authorrefmark{6, 9},
    \uppercase{Natalia Debandi}\authorrefmark{12},
    \uppercase{Agustín Gravano}\authorrefmark{4, 5, 6},
    and \uppercase{Viviana Cotik}\authorrefmark{1, 3}
}
\address[1]{Instituto de Ciencias de la Computación, CONICET, UBA (e-mail: \{jmperez, vcotik\} at dc.uba.ar)}
\address[2]{Facultad de Astronomía, Matemática y Física, Universidad Nacional de Córdoba (email: francolq at famaf.unc.edu.ar)}
\address[3]{Departamento de Computación, Facultad de Ciencias Exactas y Naturales, Universidad de Buenos Aires}
\address[4]{Laboratorio de Inteligencia Artificial, Universidad Torcuato Di Tella (email: agravano at utdt.edu)}
\address[5]{Escuela de Negocios, Universidad Torcuato Di Tella}
\address[6]{Consejo Nacional de Investigaciones Científicas y Técnicas (CONICET)}
\address[7]{Facultad de Derecho, Universidad de Buenos Aires (email: dzayat at derecho.uba.ar)}
\address[8]{Universidad Nacional del Centro de la Provincia de Buenos Aires (email: agustin.moro at azul.der.unicen.edu.ar)}
\address[9]{Instituto de Investigaciones Gino Germani, Facultad de Ciencias Sociales, Universidad de Buenos Aires}
\address[10]{Facultad de Filosofía y Letras, Universidad de Buenos Aires}
\address[11]{Escuela Interdisciplinaria de Altos Estudios Sociales, Universidad de San Martín}
\address[12]{Universidad Nacional de Río Negro (email: nataliadebandi at gmail.com)}

%% file: src/00_abstract.tex
In recent years, hate speech has gained relevance in social networks and other digital media due to its intensity and its association with violent acts against members of protected groups. Facing huge amounts of user-generated contents, a great effort has been made to develop automatic tools to aid the analysis and moderation of this kind of speech, at least in its most threatening forms. One of the limitations for current approaches on automatic hate speech detection is the lack of context. The focus on isolated messages, without considering any type of conversational context or even the topic being discussed, severely restricts the available information in orther to determine whether a post in a social network should be tagged as hateful or not. In this work, we assess the impact of adding contextual information to the hate speech detection task. In particular, we study a Twitter subdomain consisting of replies to posts by digital newspapers and media outlets, which provides a natural environment for contextualized hate speech detection. We built an original corpus in "Rioplatense" Spanish dialect focused on hate speech associated with the COVID-19 pandemic. A sample of this corpus was manually annotated  using carefully designed guidelines. Our classification experiments using state-of-the-art transformer-based machine learning techniques show evidence that adding contextual information improves the performance of hate speech detection for two proposed tasks: binary and multi-label prediction, increasing their Macro F1 by 4.2 and 5.5 points, respectively. These results highlight the importance of the use of contextual information in hate speech detection. Our code, models, and corpus has been made available for further research.

%% file: src/01_intro.tex
Hate speech can be described as speech containing denigration and violence towards an individual or a group of individuals, based on certain characteristics protected by international treaties, such as gender, race, language, and others \cite{article192015}. In recent years, this type of discourse has taken on great relevance due to its intensity and its prevalence on social media. The exposition to this phenomenon has been related to stress and depression in the victims \cite{saha2019prevalence}, and also to the setting of a hostile and dehumanizing ground for immigrants, sexual and religious minorities, as well as other vulnerable groups \cite{bilewicz2020hate}. Adding to the psychological effects, one of the most worrying aspects of hate speech on social media is its relationship with violent acts against members of these groups, such as the ``Unite the Right'' attacks at Charlottesville \cite{blout2020white}, the Pittsburgh synagogue shooting \cite{mcilroy2019welcome}, and the Rohingya genocide at Myanmar \cite{warofka2018independent,irrawaddy2018zuckerberg}, among others. As a result, states and supranational organizations such as the European Union have enacted legislation that urges social media companies to moderate and eliminate discriminatory content, with a particular focus on that which encourages physical violence \cite{european2016eu}.

The last two years have seen a dramatic increase in the prevalence of hate speech amid the COVID-19 pandemic, featuring targets such as Chinese, Asian and \hl{Jews}
, among other nationalities and minorities, blaming them for the spread of the virus or the increase in inequalities \cite{unreport2020hatespeech}. The dissemination of fake news related to conspiracy theories and other types of disinformation \cite{andersen2020proximal,cohen2020scientists} has been linked to an increase in violence against members of these groups \cite{unreport2020hatespeech}.

\newcommand\hlcite[1]{\hl{\cite{#1}}}

Great effort has been made in recent years in the research and development of automatic tools to aid the analysis and moderation of hate speech, at least in its most threatening forms \hlcite{waseem2016hateful,davidson17,schmidt2017survey,fortuna2018survey}. From a Natural Language Processing \hl{(NLP)} perspective, hate speech detection can be thought of as a text classification task: given a document generated by a user (i.e., a post in a social network), predict whether or not it contains hateful content \hlcite{schmidt2017survey}. Additionally, it may be of interest to predict other features, such as whether the text contains a call to take some possibly violent action, whether it is directed against an individual or a group, or which characteristics are  attacked \hlcite{hateval2019semeval}, for example.

One of the limitations of the current approaches to automatic hate speech detection is the lack of context. Most studies and resources work with data without any kind of context - i.e., isolated user messages with no information about the conversational thread or even the topic being discussed\hl- \hlcite{poletto2021resources}. This limits the available information to discern if a comment is hateful or not
\hl{, given that }an expression can be injurious in certain contexts, but not in others.

Another limitation is that most resources for hate speech detection are built in English, restricting the research and applicability to other languages \hlcite{schmidt2017survey,fortuna2018survey}. While there are some datasets in Spanish \hlcite{hateval2019semeval,aragon2019overview,fersini2018overview}, to the best of our knowledge, none is related to the COVID-19 pandemic, which shows distinctive features and targets in comparison to other hate speech events. \hl{Besides, none of the existing datasets comes from the Rioplatense dialectal variety of Spanish, which has its own particularities and might express hate speech in a distinct way}.


\hl{In the present work, we address the issues described above regarding hate speech detection: 1) we consider \textbf{finer-grained} distinctions that go beyond a binary detection of hateful vs. non-hateful speech, such as the identification of attacked characteristics and the detection of calls to action; 2) we study the impact of adding \textbf{contextual information} to the classification problems, and 3) we approach the problem in \textbf{Spanish}, a language with relatively few resources available for this task. We are especially interested in the second issue, regarding the usefulness of contextual information; this is the main research question of this work.}

\begin{figure*}[t]
      \centering

      \includegraphics[width=\textwidth]{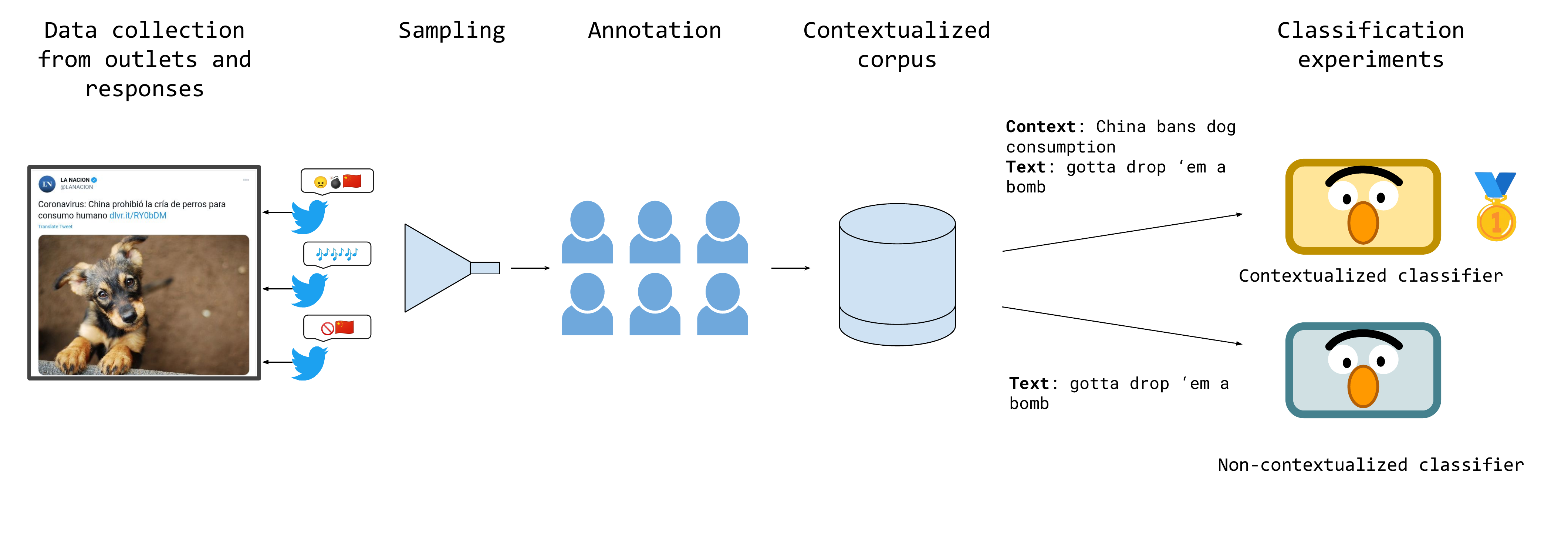}
      \caption{\hl{Work overview. The process starts with the collection of data from Twitter, according to a sampling procedure destined to achieve a balanced proportion of attacked characteristics.
                  The dataset is later annotated by native speakers following carefully designed annotation guidelines. The annotated corpus is used to train and evaluate models for hate speech detection, both as a binary and a multi-label classification task. Our experiments reveal that contextualized models outperform non-contextualized ones.
            }}
      \label{fig:process_model}
\end{figure*}

\hl{For these purposes,} we built a dataset based on user responses to posts from media outlets on Twitter. \hl{This subdomain of social networks (i.e., responses to news posts) is particularly interesting because it provides a natural context for the discussion (the news post under debate) while also replicating the interactions of a news forum.} We collected a Spanish dataset of news related to the COVID-19 pandemic and had it annotated by native speakers. Classification experiments using state-of-the-art techniques \hl{based on \textit{BETO} \cite{canete2020spanish}, a Spanish version of BERT \hl{(Bidirectional Encoder Representations from Transformers)} \cite{devlin2018bert}, show evidence that adding context improves detection both in a binary setting (predicting the presence or absence of hate speech) and in a fine-grained setting (predicting the attacked characteristics and whether there is a call to action). These results highlight the importance of contextual information for hate speech detection.}  \hl{Figure~\ref{fig:process_model} provides a graphical, high-level overview of the work discussed in this paper.}

Our contributions are the following:

\begin{enumerate}
      \item We describe the collection, curation and annotation process of a novel corpus for hate speech detection based on user responses to news posts from media outlets on Twitter. This dataset is in the Rioplatense dialectal variety of Spanish and focuses on hate speech associated with the COVID-19 pandemic.
      \item \hl{Through a series of classification experiments using state-of-the-art techniques, we show evidence that including contextual information improves the performance of hate speech detection, both in binary and fine-grained settings.}
      \item We make our code, models and the annotated corpus available for further research.\footnote{Our code and corpus will be publicly available once the paper is published. If needed before, please write to the corresponding author.}
\end{enumerate}

The rest of the paper is organized as follows: Section~\ref{sec:previous_work} reviews previous work for automatic hate speech detection. Section~\ref{sec:hate_speech_definition} states the definition of hate speech used in this work, along with the targeted groups and the characteristics of interest. Section~\ref{sec:corpus} describes the process performed to collect and annotate our corpus, which is later used in Section~\ref{sec:classification_experiments} to conduct our classification experiments. Section~\ref{sec:discussion} discusses the results and Section~\ref{sec:conclusions} draws conclusions and outlines possible future work.


%% file: src/02_previous.tex
\label{sec:previous_work}

Hate speech has attracted a lot of attention in recent years, with literature from the legal and social domains studying its definition and classification \cite{cele2019}, the elements that enable its identification, and its relationship to freedom of expression and human rights  \hl{\cite{article192015, CIDH2015}}. The automatic detection of this phenomenon is usually approached as a classification task, and is related to a family of other tasks such as cyberbullying, offensive language, abusive language, toxic language, and others. \citet{waseem-etal-2017-understanding} propose a typology of these related tasks by asking whether the offensive content is directed to a specific entity or group, and whether the content is explicit or implicit.

There is a plethora of resources for the automatic detection of hate speech\hl{.  I}nterested readers can refer to \citet{poletto2021resources} for an extensive review of datasets for this task. In particular, Spanish corpora are scarce, despite its being one of the most used languages in social media, and the second language in the number of native speakers worldwide
\cite{ethnologue}. To the best of our knowledge, all available datasets for this language have been published in the context of shared tasks. \citet{fersini2018overview} presented a $\sim$4k Twitter dataset for the Automatic Misogyny Identification (AMI) shared task (IberEval 2018\footnote{\hl{IberEval 2018: \url{https://sites.google.com/view/ibereval-2018?pli=1}}}). The MEX-A3T task (IberEval 2018 and IberLEF 2019\footnote{\hl{IberLEF 2019: \url{https://sites.google.com/view/iberlef-2019/}}}) included a dataset of $\sim$11k Mexican Spanish tweets annotated for aggressiveness \cite{mex-a3t2018,mex-a3t2019}. \citet{hateval2019semeval} published a $\sim$6.6k tweets dataset annotated for misogyny and xenophobia, in the context of the HatEval challenge (SemEval 2019 \footnote{\hl{SemEval 2019: \url{https://alt.qcri.org/semeval2019/}}}).

Due to the COVID-19 pandemic, a spike in the incidence of hate speech has been documented in social networks \cite{hswen2021association}. Some works have addressed its distinctive features, studying hateful dynamics in social networks \cite{uyheng2021characterizing} and also generating specific resources for the analysis and identification of this kind of toxic behavior \cite{li2021covid}.
\citet{cotik2020studyBlind} describe a work-in-progress on this research of hate speech in Spanish tweets related to newspaper articles about the COVID-19 pandemic.

Regarding techniques for our specific task, classic machine learning techniques such as handcrafted features and bags of words over linear classifiers have been applied \cite{waseem2016hateful,greevy2004classifying,warner2012detecting}. Lately, however, deep learning techniques such as recurrent neural networks or ---more recently--- pre-trained language models have become state-of-the-art  \cite{gamback17,park17,badjatiya2017deep,agrawal18,bisht2020, perez2019atalaya}. In spite of the great results achieved by these methods, \citet{arango2019} calls some of them into question, suggesting that they may be due to possible cases of overfitting. \citet{plaza2021comparing} analyze the currently available Spanish pre-trained models for hate speech detection tasks.

Since the appearance of GPT \hl{(Generative Pre-trained Transformer)} \cite{radford2018improving} and BERT \cite{devlin2018bert}, pre-trained language models based on transformers \cite{vaswani2017attention} have become state-of-the-art for most NLP tasks. \hl{These techniques use a transfer-learning approach, by first pre-training a large language model (thus their name) on a big corpus, and then fine-tuning it for a specific task (e.g. sentiment analysis, question answering, or hate speech detection) \cite{radford2018improving,howard-ruder-2018-universal}. This approach has replaced previous deep learning architectures for most NLP tasks, which used to be based on recurrent neural networks and word embeddings \cite{iyyer2014neural,huang2015bidirectional}.}

\hl{Pre-trained models have been built for different languages, and also for different domains (such as the biomedical \cite{lee2020biobert} and the legal domains \cite{chalkidis-2020-legal}) and text sources (such as Twitter \cite{nguyen-etal-2020-bertweet} and other social networks).} In particular, Spanish pre-trained models include BETO \cite{canete2020spanish}, BERTin \cite{BERTIN}, RoBERTA-es \cite{gutierrez2022maria} and RoBERTuito \cite{perez2022robertuito}. \citet{nozza2020mask} review BERT-based language models for different tasks and languages.\footnote{\hl{Note that the names BETO, BERTin, RoBERTA and RoBERTuito are not acronyms, but alterations of the original name BERT.}}

\hl{Few prior studies incorporate some kind of context to the user comments for hate speech or toxicity detection}. \citet{gao2017detecting} analyze the impact of adding context to the task of hate speech detection for a dataset of comments from the Fox News site. As mentioned by \citet{pavlopoulos2020toxicity}, this study has room for improvement: the dataset is rather small, with around 1.6k comments extracted from only 10 news articles; its annotation process was mainly performed by just one person; and some of its methodologies are subject to discussion, such as including the name of the user as a predictive feature.
\citet{mubarak-etal-2017-abusive} built a dataset of comments taken from the Al Jazeera website,\footnote{\url{https://www.aljazeera.com/}} and annotated them together with the title of the article, but without including the entire thread of replies.

\citet{pavlopoulos2020toxicity} analyze the impact of adding context to the toxicity detection task. They find that, while humans seem to leverage conversational context to detect toxicity, the trained classification models were not able to improve their performance significantly by adding context. Following up, \citet{xenos-2021-context} label each message with its ``context sensitiveness'', measured as the difference between two groups of annotators: those who have seen the context, and those who have not. With this, they observe that classifiers improve their performance on comments which are more sensitive to context.

Further, \citet{sheth2021defining} explore some opportunities for incorporating richer information sources into the toxicity detection task, such as the interaction history between users, some kind of social context, and other external knowledge bases. \citet{wiegand2021implicitly} pose some questions and challenges regarding the detection of implicit toxicity --- that is, some subtle forms of abusive language not expressed as slurs.

\hl{Summing up, BERT-based models are state-of-the-art for this type of classification tasks; there have been various attempts to include context in distinct ways and with disparate success; there have been relatively few studies on Spanish data; and hate speech detection has typically been addressed as a binary task, making no distinction among the attacked characteristics or calls-to-action. In the present work, we assess the usefulness of adding context, we work with BERT-based models, on Spanish data, and address both binary and fine-grained classification tasks.}

%% file: src/03_hate_speech_definition.tex
\label{sec:hate_speech_definition}

We say that there is hate speech in a comment if it contains statements of an intense and irrational nature of disapproval and hatred against an individual or a group of people because of its identification with a group protected by domestic or international laws \cite{article192015}. Protected treats or characteristics include color, race, national or social origin, gender identity, language, and sexual orientation, among others.

Hate speech can manifest itself explicitly as direct insults, slurs, celebrations of crimes, incitements to take action against an individual or group, or even more veiled expressions such as ironic content. Following this definition, we consider that an insult or aggression is not enough to constitute hate speech; it is necessary to make an explicit or implicit appeal to at least one protected characteristic.

For international law, hate speech has an extra element that differentiates it from offensive behavior: the promotion of violent actions against its targets. However, the NLP community does not usually require this ``call to action'' when identifying hate speech. In the present work, we will adopt this latter view, and we will explicitly state when we also refer to calls to action.

Several characteristics are taken into account in this work. In addition to misogyny and racism (the most common treats considered in previous work\hl{s}), we also consider: homophobia and transphobia; social class hatred (sometimes known as aporophobia); hatred due to physical appearance (e.g., overweight); hatred towards people with disabilities; political hate speech; and hate speech against criminals, prisoners, offenders and other people in conflict with the law. For this selection, we take into account the definition of discrimination from international human rights treaties, which refers to discrimination motivated by race, color, sex, language, religion, political, or other opinions, national or social origin, property, birth or other status \cite{generalcomment20}.
These eight characteristics are listed in Table \ref{tab:protected_characteristics} along with reference names that will be used throughout the paper.

\begin{table}[]
    \centering
    \begin{tabular}{l p{0.33\textwidth}}
        Short name & Hate speech against ...                                                     \\
        \hline
        WOMEN      & women                                                                       \\
        LGBTI      & gay, lesbian, bisexual, transgender, intersexual people                     \\
        RACISM     & people based on their race, \hl{skin} color, language, or national identity \\
        CLASS      & people based on their socioeconomic status                                  \\
        POLITICS   & people based on their political affiliation or ideology                     \\
        APPEARANCE & fat people, old people, or other aspect-based features                      \\
        CRIMINAL   & criminals and persons in conflict with law                                  \\
        DISABLED   & people with disability or mental health affections                          \\
        \hline
    \end{tabular}
    \caption{Protected characteristics considered in this work. Short names are used throughout the paper to refer to these broad groups.}
    \label{tab:protected_characteristics}
\end{table}

%% file: src/04_corpus.tex
\label{sec:corpus}

\begin{figure*}[ht]
    \centering
    \includegraphics[width=0.90\textwidth]{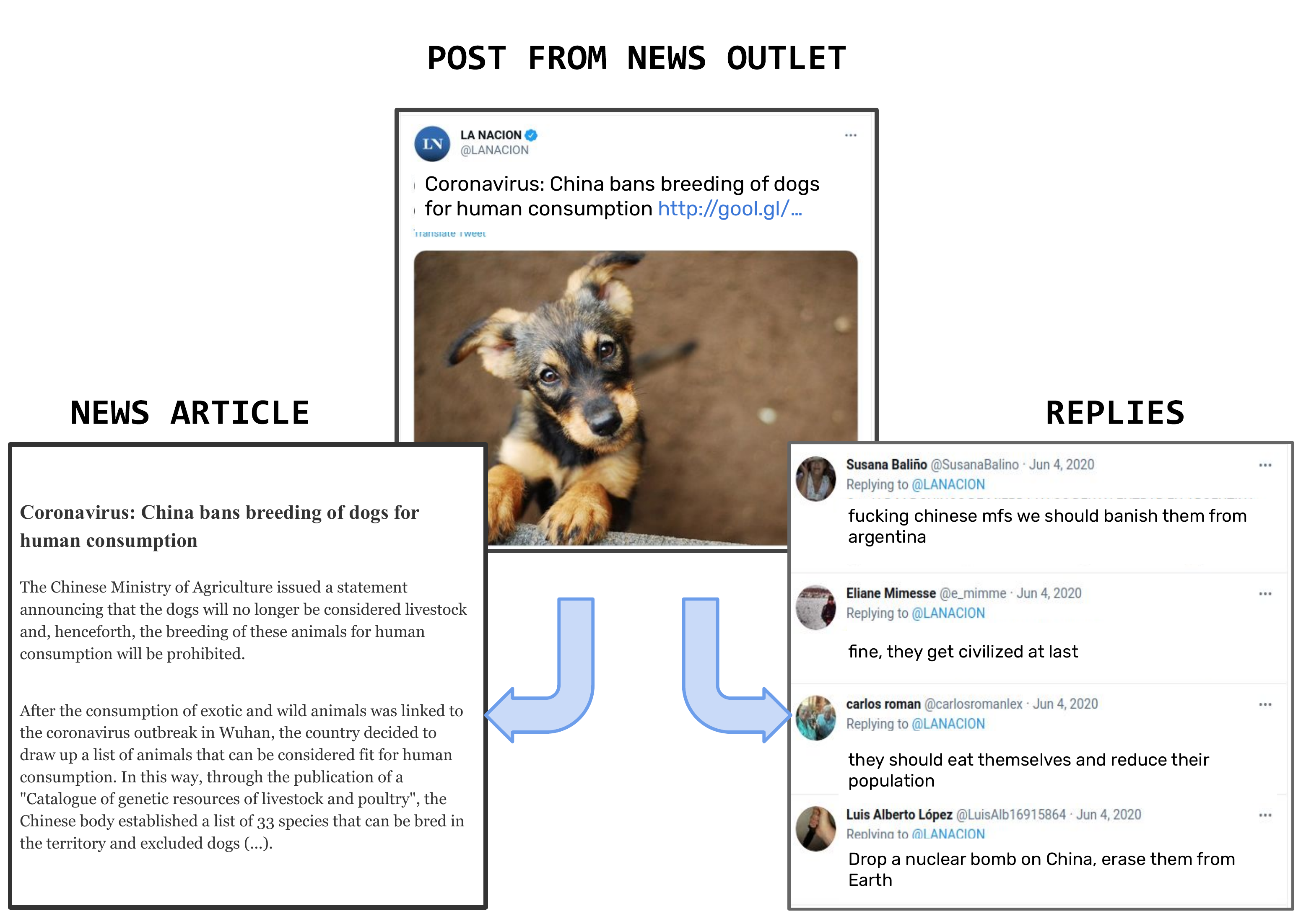}
    \caption{\hl{Example illustrating the elements in our corpus: a news article (bottom left), a tweet referring to it (top), and its replies from Twitter users (bottom right). The user comments are the instances analyzed as potential hate speech; the original tweet and the article itself are the context. All texts in this Figure were translated from Spanish to English.}}
    \label{fig:idea_dataset}
\end{figure*}

This section describes the collection, curation, and annotation process of the corpus. Our aim was to construct a dataset of user messages commenting on specific news articles, in a similar fashion to the reader forums present in many news outlet websites.
Figure \ref{fig:idea_dataset} offers a schematic illustration of our dataset, with a tweet from a news outlet about China banning the breeding of dogs for human consumption, its respective news article, and replies from users to the original tweet.

\subsection{Data Collection}
\label{sec:data_collection}

Our data collection process was targeted at the official Twitter accounts of a selected set of Argentinian news outlets: La~Nación (@lanacion), Clarín (@clarincom), Infobae (@infobae), Perfil (@perfilcom), and Crónica (@cronica). These are the main National newspapers in the country, and attract a vast volume of interaction on Twitter.

We considered a fixed time period of one year, starting in March 2020. We collected the replies to each post of the mentioned accounts using the \emph{Spritzer} Twitter API, listening to any tweet mentioning one of their usernames.

For the purpose of this work, we were only interested in the first level of replies to the original tweet, in order to consider as context only the news under discussion. If the second or further levels of replies had been considered, the context would have also contained comments made by other users (i.e., a conversational thread), which we wanted to avoid. Also, we discarded tweets from outlets that were not linked to a news article.

To focus our dataset on hate speech related to the COVID-19 pandemic, we only kept those articles whose body contained one of the following terms: coronavirus, COVID-19, COVID, Wuhan, \emph{cuarentena} (quarantine),  \emph{normalidad} (normality), \emph{aislamiento} (isolation), \emph{padecimiento} (suffering), \emph{encierro} (confinement),  \emph{fase} (phase), \emph{infectado} (infected), \emph{distanciamiento} (distancing), \emph{fiebre} (fever) and \emph{síntoma} (symptom).

Hate speech is not evenly distributed across news articles or topics of discussion. Previous work has focused on multiple strategies to detect users or topics around which this phenomenon is prevalent: for example, monitoring specific targets, hashtags, or offending users \cite{hateval2019semeval}. In this case, some form of sampling strategy is also necessary before the annotation step, since a random sample of the collected data would bring a very small proportion of hateful messages.

One of the sampling strategies we considered was to use some keywords to select interesting articles, taking into account topics that could be a focus of hate speech. The second strategy considered was to sample articles based on their comments: news containing comments with common slurs or pejorative expressions towards our protected groups. That is,  we kept only news articles containing two or more comments that are marked according to a list of predefined slurs. We selected expressions and slurs that addressed the protected characteristics considered in our hate speech definition, described in Section \ref{sec:hate_speech_definition}. The list of slurs and some other technical details are described in Appendix \ref{app:data-selection}.

After some experimentation and subjective evaluation of the articles retrieved using each strategy, we decided to use the latter one --- i.e., to select news articles based on their user comments --- as it seemed to produce the best results. We emphasize that we sampled the whole article and its comments, and not just the replies that contained slurs. For each sampled article, 50 comments were chosen at random for annotation, after excluding those with URLs or images.

Finally, we anonymized tweets by removing user handles and replacing them with a special \texttt{@user} token, as there are some accounts usually mentioned by hateful users that could bias the annotation process.

\subsection{Annotators}
\label{sec:annotators}

Considering that hate speech is usually manifested through jargon and slurs, and with a strong socio-cultural background, we hired six native speakers of the Rioplatense dialect of Spanish. Following the lines of Data Statements \cite{bender2018data}, we provide in this subsection a characterization of the annotators.

The expected profile of the annotators was of students or graduates of social sciences, humanities, or related careers, with no experience in artificial intelligence or data science (to avoid biases in the task). It was also of interest that they were frequent users of social networks so they could capture the subtleties of language in that medium.

As part of the recruitment process, they were asked to take a paid test that consisted in reading the guidelines and annotating ten articles with their respective comments. After evaluating the results of this test, no applicants were rejected in this step.

Table \ref{tab:information_about_notators} provides disaggregated information about the six annotators hired for the task. All six had a highly educated profile, and two of them had previous experience with labeling data. At the time of the study, two of the labelers were activists in organizations related to some of the vulnerable groups considered in this work. Four of them identified themselves as members of groups targeted by hate speech: women and LGBTI \hl{(lesbian, gay, bisexual, transgender and intersex)}.

\begin{table}[h]
    \centering
    \footnotesize
    \begin{tabular}{l l l l }
        Gender & Age   & Educ.     & Area               \\
        \hline
        F      & 25-30 & PhD*      & \hl{Psychology}    \\
        NB     & 30-35 & Undergrad & Arts               \\
        F      & 30-35 & Undergrad & \hl{Anthropology}  \\
        M      & 35-40 & Graduate  & Sociology          \\
        F      & 35-40 & PhD       & \hl{Psychology}    \\
        F      & 30-35 & Graduate  & \hl{Communication} \\
        \hline
    \end{tabular}
    \caption{Information about the annotators: gender, age \hl{range}, education, area of studies. * indicates ongoing. 
        F stands for female, M for male, NB for non-binary. }
    \label{tab:information_about_notators}
\end{table}

\subsection{Annotation process}

%
%
%
%

\begin{figure}[t]
    \centering
    \includegraphics[width=0.45\textwidth]{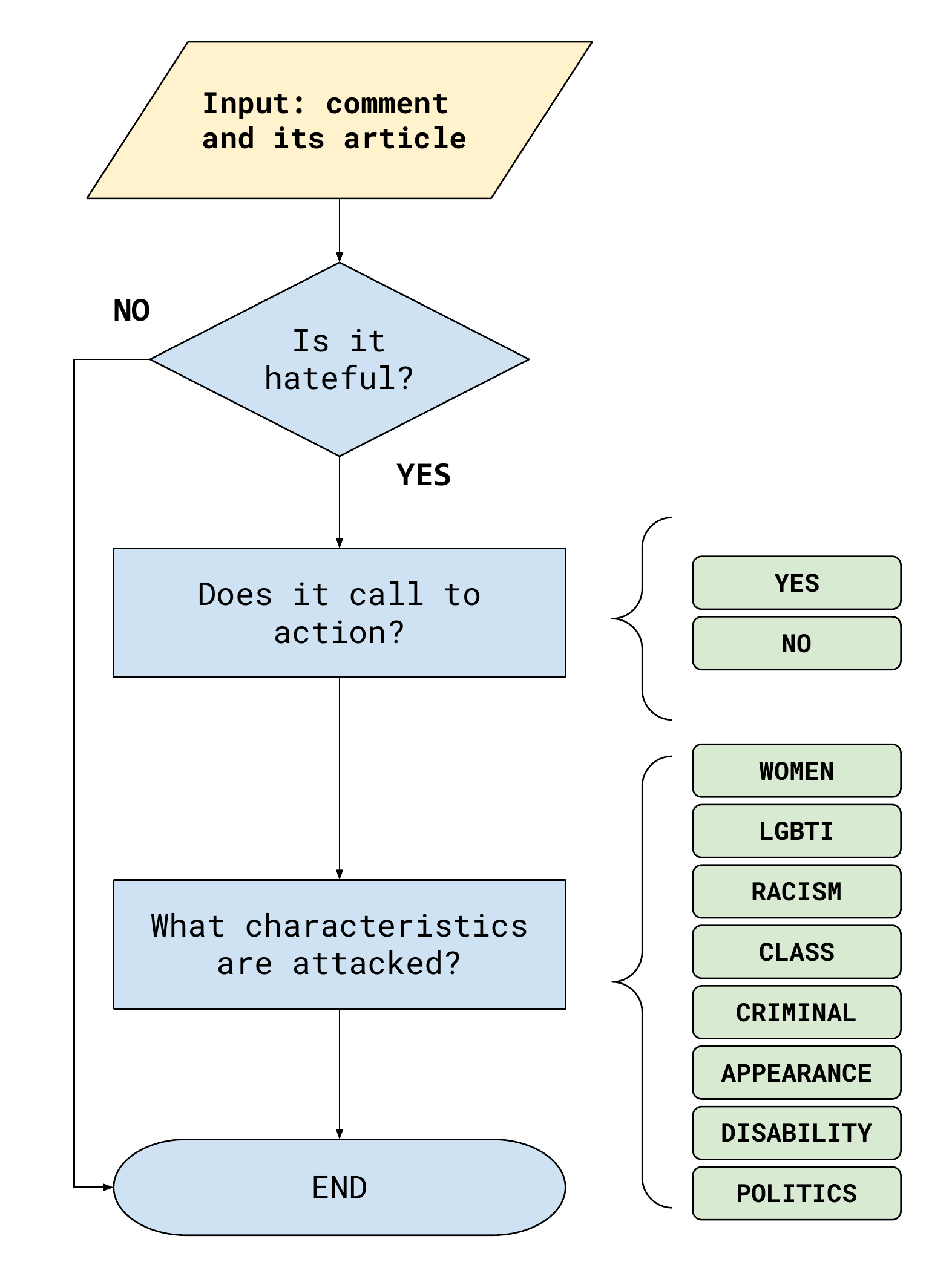}
    \caption{Annotation model for each pair of articles and comments\hl{.}}
    \label{fig:annotation_model}
\end{figure}

To annotate our data, we followed a similar process to the MAMA portion of the MATTER cycle \cite{pustejovsky2012natural}. First, we defined our model; that is, a practical representation of what we intended to annotate. Figure \ref{fig:annotation_model} depicts the model used in this work, which follows a hierarchical structure as proposed by \citet{zampieri2019predicting}. For each comment and its respective context (the tweet from the news outlet and the full article), a first annotation requires to mark whether the comment is hateful or not. If it is marked as not hateful, no further information is required. If it is marked as hateful, two extra annotations are required:

\begin{itemize}
    \item An annotation to indicate whether the comment contains a call to action; and
    \item One or more annotations for each protected characteristic that is attacked by the message.
\end{itemize}

Each annotation task comprised a newspaper article along with each of the selected comments for it. Annotators were given the option of skipping an article when they considered it irrelevant in terms of hate speech, or when they did not want to annotate it due to personal reasons (no one actually skipped an article due to this).

For each article, up to 50 comments were displayed. The annotator had to label the comments following the hierarchical schema shown in Figure \ref{fig:annotation_model}. Each article was first presented to two different annotators with all its comments. Then, a third annotator only had to annotate those comments marked by at least one of the previous workers as hateful. While for a majority voting scheme it would just be necessary to check those with exactly one hateful annotation, an extra annotation was collected for further experiments.

Before beginning their task, each annotator was required to go through a training period, which consisted of the test mentioned in Section \ref{sec:annotators} and a second step of annotating 15 articles. This was the only set of articles labeled by all the annotators. At the end of this stage, they were given feedback to adjust their criteria, and then proceeded to the actual annotation task.

\subsection{Dataset results}
\label{sec:dataset_results}

\begin{table}[b]
    \centering

    \begin{tabular}{lrrr}
        \toprule
        Characteristic & Count      & Calls to action & $\alpha$     \\
        \hline
        RACISM         & \num{2469} & 674             & \hl{$0.608$} \\
        APPEARANCE     & \num{1803} & 34              & \hl{$0.735$} \\
        CRIMINAL       & \num{1642} & 722             & \hl{$0.618$} \\
        POLITICS       & \num{1428} & 136             & \hl{$0.509$} \\
        WOMEN          & \num{1332} & 18              & \hl{$0.531$} \\
        CLASS          & \num{ 823} & 135             & \hl{$0.404$} \\
        LGBTI          & \num{ 818} & 11              & \hl{$0.555$} \\
        DISABLED       & \num{ 580} & 4               & \hl{$0.596$} \\
        \hline
    \end{tabular}
    \caption{Figures of hateful tweets in our dataset (i.e. annotated by at least two annotators as hateful), segmented by characteristic with the corresponding number of tweets calling to action. \hl{Inter-annotator agreement is given for each characteristic, as measured by Krippendorff's alpha.}}
    \label{tab:dataset_figures}
\end{table}

The resulting dataset consists of \num{56869} tweets from \num{1238} news articles. From these tweets, \num{8715} tweets were marked as hateful by two or three annotators. Table \ref{tab:dataset_figures} displays the number of hateful tweets for each of the considered characteristics. The predominant class of hateful tweets corresponds to racism, followed by tweets offending by appearance.

Calls to action are mainly directed against criminals, and also driven by racist motives. Hateful tweets due to class and political reasons have some tweets in this category as well, and the other characteristics do not account for much of these violent interactions. Table \ref{tab:examples} displays some examples of hateful tweets with their corresponding annotations.

\input{src/examples.tex}

From the \num{8715} hateful comments, 77\% (\num{6777}) contain only one attacked characteristic, nearly 20\% have two or more, and 220 comments have three or more. Maximum co-occurrence occurs between the characteristics WOMEN and APPEARANCE, followed by RACISM and CLASS, POLITICS and CLASS, and RACISM and POLITICS. More information about the co-occurrence of attacked characteristics can be found in Appendix \ref{app:additional_information}.

As suggested by \citet{arango2019}, we checked the distribution of users generating hateful content, so as to avoid having a small number of users responsible for the majority of offensive interactions. The mean amount of hateful comments per user is $1.44$, with only 28 users having more than 10 hateful comments.

Inter-annotator agreement was measured via
Krippendorff's alpha \cite{krippendorff2011computing}, using the implementation included in the \verb|krippendorff| library for Python.\footnote{\url{https://github.com/pln-fing-udelar/fast-krippendorff}} The agreement for the hate speech label was $0.579$, which is compatible with other studies in the area, and expectable considering that we used a rather broad definition of hate speech \cite{poletto2021resources}. For the \emph{calls-to-action} label, the agreement was slighly higher at $0.641$. Individual agreements for each characteristic are displayed in Table \ref{tab:dataset_figures}.

To assign gold labels for each tweet in the dataset, we followed a majority-vote strategy. A tweet was marked as hateful if at least two annotators labeled it as such. The CALLS label (calls-to-action) was marked if at least two annotators selected it, and we marked each characteristic if at least one annotator selected it. When a tweet was not marked as hateful, no other labels were assigned.

%% file: src/examples.tex
\begin{table*}[t]
    \centering
    \small
    \begin{tabular}{l p{0.4\textwidth} p{0.4\textwidth}}
        \toprule
        Characteristic          & Context                                                                                                  & Comment                                                                                                                                                                                  \\
        \midrule
        WOMEN                   & Around the world: Florencia Peña shows her luxurious new house with bar, dock and pool                   & @usuario When you suck the right ones                                                                                                                                                    \\
        \rule{0pt}{3ex}
        \rule{0pt}{3ex}WOMEN    & Mia Khalifa: acted in porn videos for a few months, became world famous and now fights to erase her past & @usuario HAHAHAHAHAHAHA KEEP SUCKING....                                                                                                                                                 \\
        \rule{0pt}{3ex}WOMEN    & Narda Lepes: ``They touched my ass a thousand times in restaurant kitchens''                             & @user Do you have a nice ass?                                                                                                                                                            \\
        \midrule
        LGBTI                   & Why Flor de la V did not continue in Mujeres de eltrece, after the departure of Claudia Fontán           & @usuario ...because she is not a woman, crystal clear                                                                                                                                    \\
        \rule{0pt}{3ex}LGBTI    & Historical: Mara Gómez was enabled and will be the first trans player in Argentine soccer                & @usuario What pair of balls this girl has!!!                                                                                                                                             \\
        \rule{0pt}{3ex}LGBTI    & The story of the Colombian trans model kissing the belly of her eight-month pregnant husband             & @usuario A male kissing another male                                                                                                                                                     \\
        \rule{0pt}{3ex}LGBTI    & This is what actor Elliot Page looks like after declaring himself trans                                  & @user she has bick\footnote{We translated ``bija'' (a purposely misspelling of ``pija'') as ``bick''}? No. she has pussy? Yes. She is a woman                                            \\
        \midrule

        RACISM                  & Coronavirus. Yanzhong Huang: "It is quite likely that a Covid-21 is already brewing"                     & @user Urgent bombs to that damned race                                                                                                                                                   \\
        \rule{0pt}{3ex}RACISM   & Scientists denounced China's new maneuver to hide the true figures of the coronavirus                    & @user Globally we maintain China because everything comes from there and today we are melted and in an emergency... \#ChinaVirus I don't want to see a \#Chinese for a long time!        \\
        \rule{0pt}{3ex}RACISM   & Impressive operation with tanks for a prosecutor to enter an area controlled by Mapuches                 & @usuario Stop it!!! They are not Mapuches, they are criminals!!! Let's see if someone puts the balls where they have to be put and they shoot them down at once!                         \\

        \midrule
        CRIMINAL                & Rosario: a group of neighbors beats to death a young man accused of stealing cars                        & @user this is great, an example to others                                                                                                                                                \\
        \rule{0pt}{3ex}CRIMINAL & A guy takes the gun from the thief who assaulted him, runs him off and shoots him dead: arrested         & @usuario Great, let's go for the total extermination of these apes.                                                                                                                      \\

        \midrule
        \rule{0pt}{3ex} CLASS   & Social movements cut off 9 de Julio Av.: they demand a minimum wage of \$45,000                          & @user get to work, mfs.                                                                                                                                                                  \\                                                                                                                              \\
        \rule{0pt}{3ex}POLITICS & A new COVID-19 mutation is confirmed, up to 10 times more contagious than the original strain from Wuhan & @usuario I'M VERY GLAD. I HOPE IT WILL ARRIVE SOON IN ARGENTINA AND DESTROY EVERYTHING. WE COULD FINALLY SEE SOMETHING MORE HARMFUL THAN PERONIST CANCER AND ITS KIRCHNERIST METASTASIS. \\

        \hline
    \end{tabular}
    \caption{Some hateful examples of our dataset for each of the considered characteristics.}
    \label{tab:examples}
\end{table*}

%% file: src/05_classifiers.tex
\label{sec:classification_experiments}

\begin{figure}[t]
    \centering
    \includegraphics[width=0.5\textwidth]{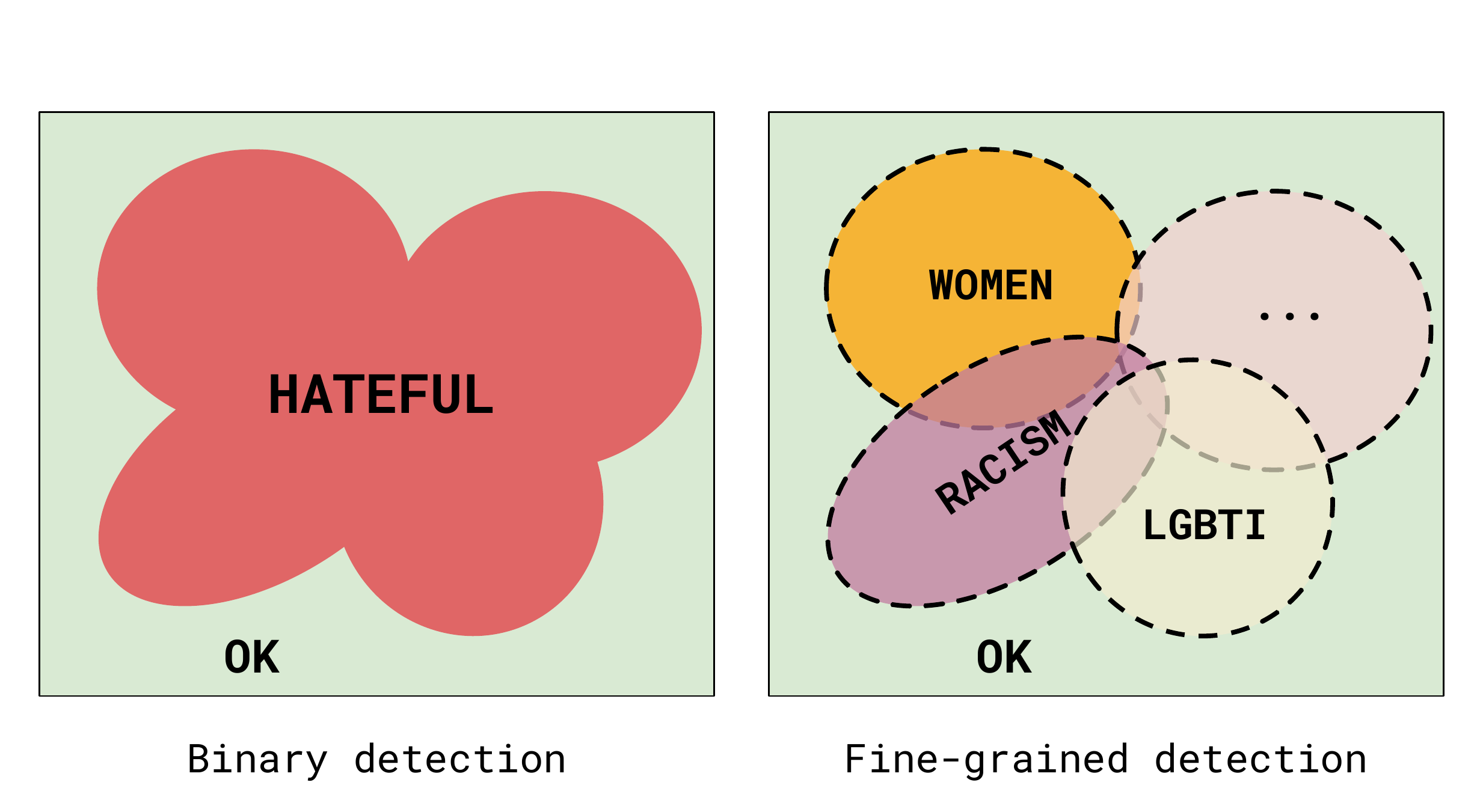}
    \caption{Proposed tasks. \hl{The binary task consists in predicting whether a tweet is hateful or not. The fine-grained task consists in predicting the attacked characteristics, and whether it calls to action or not.}}
    \label{fig:hate_detection_tasks}
\end{figure}

Now that we have this specially-crafted corpus containing context, we turn our attention to our original research question: can classifiers leverage context to improve their performance on the hate speech detection task?
For this purpose, we propose the following classification tasks:

\begin{itemize}
    \item \textbf{Binary} hate speech detection: Given a tweet, predict whether it is hateful or not.
    \item \textbf{Fine-grained} hate speech detection: Given a tweet, predict the attacked characteristics (if any), and whether it calls to action or not.
\end{itemize}

In machine learning terms, the binary task can be posed as a binary classification task, while the fine-grained task is a multi-label classification task. Figure \ref{fig:hate_detection_tasks} illustrates the difference between both tasks as a Venn diagram: in the binary task, we have to predict whether a tweet belongs to the set of hateful tweets; whereas in the fine-grained one, we have to predict if a tweet belongs to the set of hateful tweets for each given characteristic (eight, in our case). The binary task can be seen as a simpler form of the fine-grained task.

\subsection{Classification algorithms}
\label{sec:classification_algorithms}



\begin{figure*}[t]
    \centering
    \begin{subfigure}[]{0.45\textwidth}
        \includegraphics[width=\textwidth]{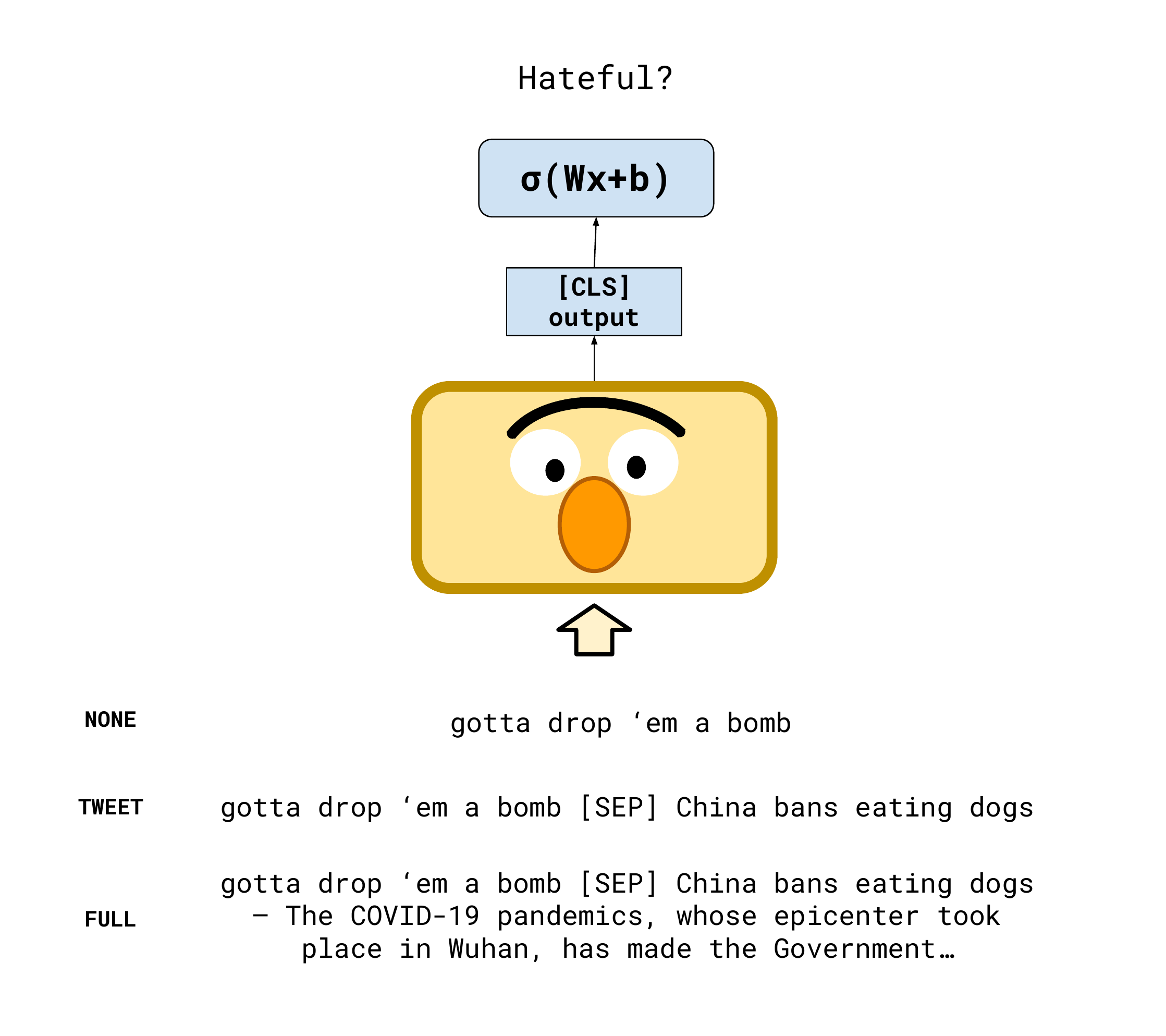}
        \caption{Binary classifier}
        \label{subfig:binary_classifier}
    \end{subfigure}
    \begin{subfigure}[]{0.45\textwidth}
        \centering
        \includegraphics[width=\textwidth]{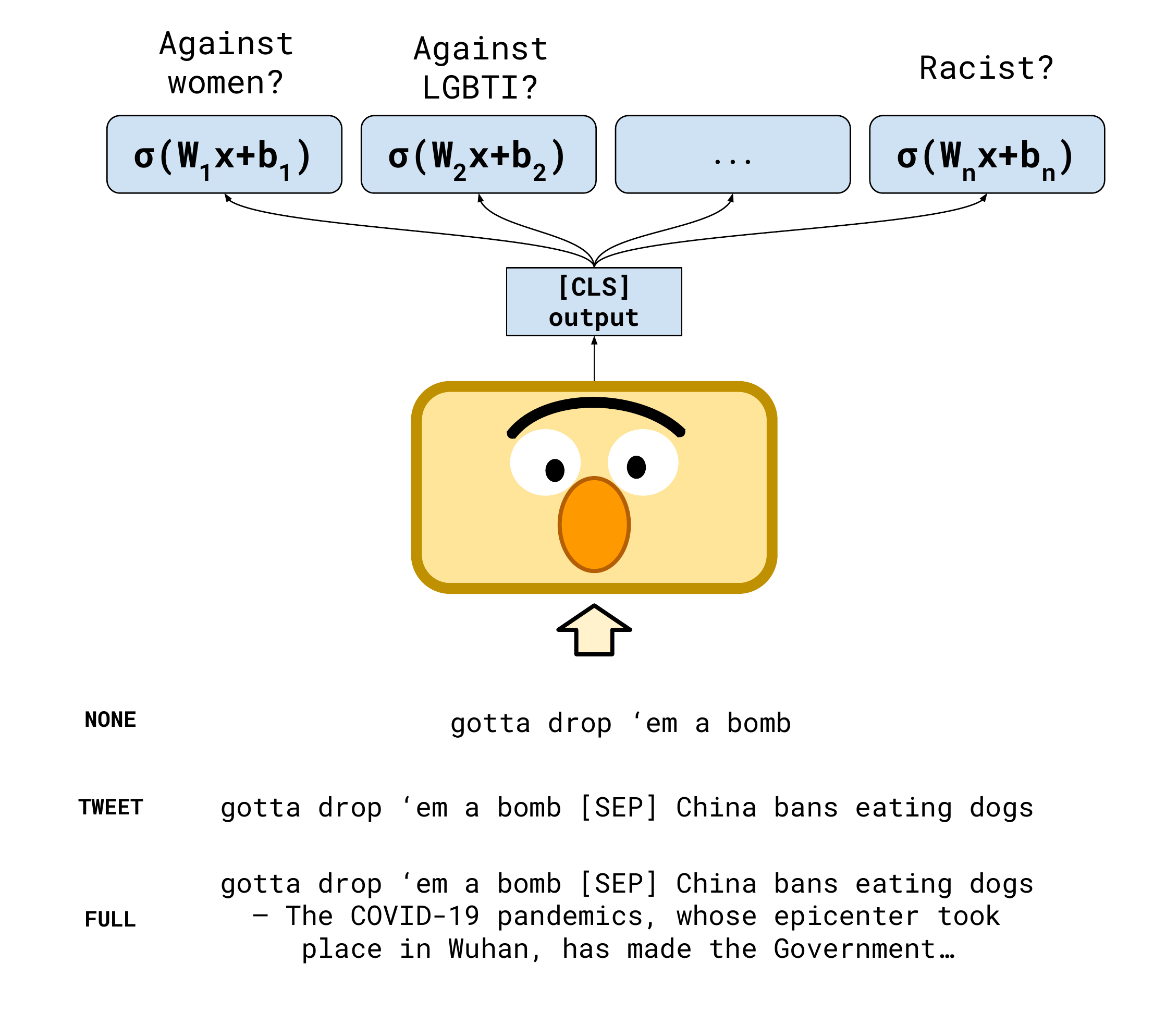}
        \caption{Fine-grained classifier}
        \label{subfig:fine_grained_classifier}
    \end{subfigure}

    \caption{\hl{Classification models for the proposed tasks. Three different types of classifiers are trained according to the type of input: \textbf{None} (no context), \textbf{Tweet} (context is the tweet to which the comment responds), and \textbf{Full} (context is the tweet to which the comment responds plus the text of the news article).}}
    \label{fig:classification_models}
\end{figure*}

For both tasks, we trained algorithms based on state-of-the-art classification techniques, namely BERT. \hl{As explained in Section \ref{sec:previous_work}, BERT models are based on Transformer language models pre-trained on large corpora. To adapt them for a specific task, a fine-tuning process is performed, which consists in removing the last layer of the language model (usually, a big softmax for the Cloze task\footnote{\hl{The Cloze task is a widely used task to evaluate an NLP system's language understanding ability, that consists in replacing a missing part of a text.}}), replacing it with a proper layer for the downstream task (e.g. sentiment analysis, question answering), and then adjusting the weights of the whole model \cite{devlin2018bert,radford2018improving}.  }

\hl{
    Since our dataset is in Spanish, we used BETO \cite{canete2020spanish}, a monolingual BERT model for this language. We employed its base version, which consists of 12 Transformer layers with 12 attention heads each, summing up around 100M parameters.}

To assess the importance of having contextual information, we considered three different types of inputs for the proposed models: the comment without any context (which we call \textbf{None}), the comment with the tweet to which it responds as context (\textbf{Tweet}), and the comment with the tweet to which it responds plus the text of the news article (\textbf{Full}). The special \texttt{[SEP]} token is used to encode the separation between the context and the analyzed text in the \textbf{Tweet} and \textbf{Full} inputs (our two context-aware models).

\hl{For the binary task, we trained a standard BERT architecture for binary sequence classification \cite{devlin2018bert}, consisting of a sigmoidal output consuming the last hidden state of the \texttt{[CLS]} token, which acts as a continuous representation for the whole sentence. For the fine-grained task, we propose a multi-label output; that is, the simultaneous prediction of the eight characteristics and the call-to-action label. Figure \ref{fig:classification_models} illustrates both models for their three different types of inputs. }

\subsection{Training}

\hl{We trained the classifiers following the guidelines of \citet{devlin2018bert}.  We used Adam \cite{kingma2014adam} as the optimizer, with a weight decay of $0.1$, a peak learning rate of $5 * 10^{-5}$ (at the 10\% of the optimization steps), and a batch size of 32. We trained the model for 5 epochs, and selected the best model according to the F1 score on the dev set. The loss function for the binary detection task was the binary cross-entropy loss, defined as}

\hl{
    \begin{equation*}
        L_b(y, \widehat{y}) = -y \log(\widehat{y}) - (1 - y) \log(1 - \widehat{y})
    \end{equation*}
    \noindent where $y$ is the true label (0 or 1) and $\widehat{y}$ is the predicted probability of the positive class.
}

\hl{The training process for the fine-grained models was mostly the same, with the exception of the loss function. As the output of the model is a vector of probabilities for each output variable (eight characteristics plus call-to-action), we used a multi-label loss function that considers the probability of each class independently. Let $d$ be the number of output variables (9 in our case), $y \in \{0, 1\}^d$ the true label vector, and $\hat{y} \in [0, 1]^d$ the predicted probabilities. Then, the loss function is defined as:
    \begin{equation*}
        L(y, \widehat{y}) = \sum\limits_{i=1}^d L_b(y_i, \widehat{y}_i)
    \end{equation*}
    \noindent where $L_b$ is the binary cross-entropy previously defined}.

Sharing the weights between all of the outputs has two benefits: first, it allows for the creation of a more compact model (otherwise there would be nine different BERTs adding up to a billion parameters); and second, it enables sharing common information between the different attacked features.
\hl{Further details about the training process can be found in Appendix \ref{app:classification_experiments}.}

\subsection{Domain adaptation}

\hl{ Standard training of BERT-based classifiers includes two steps as explained in Section \ref{sec:classification_algorithms}: the pre-training of the language model and the fine-tuning of the model to the downstream task \cite{devlin2018bert}. Other transfer-learning approaches in NLP ---such as Universal Language Model Fine-tuning \cite{howard-ruder-2018-universal}--- incorporate an intermediate step, that adjusts the pre-trained model to the target domain by continuing the language modeling using the text of the downstream task. \citet{gururangan-2020-dont} showed that continuing the pre-training of BERT-based models on the target domain improves the performance of the models for several subdomains of tasks.

    In our experimental setup, we adapted BETO using a sample of comments and articles discarded from the annotation process. As we had three different types of inputs, we performed three domain adaptations according to the shape of the input, as shown in Figure \ref{fig:classification_models}.
    \begin{table}[t]
        \centering
        \small
        \begin{tabular}{lr}
            \toprule
            Hyperparameter  & Value            \\
            \midrule
            Steps           & \num{10000}      \\
            Batch size      & \num{2048}       \\
            Max Seq. Length & 128, 256 and 512 \\
            $\beta_1$       & $0.9$            \\
            $\beta_2$       & $0.98$           \\
            $\epsilon$      & $10^{-6}$        \\
            Decay           & $0.01$           \\
            Peak LR         & $0.0004$         \\
            Warmup ratio    & $0.1$            \\
            \bottomrule
        \end{tabular}
        \caption{Hyperparameters used for domain adaptation\hl{.}}
        \label{tab:hs_ft_hyperparameter}
    \end{table}
    Table \ref{tab:hs_ft_hyperparameter} contains the hyperparameters used to adapt the \beto{} model to our domain. We used the remaining data of the collection process, consisting of around \num{288000} articles and \num{5000000} comments. Three versions of \beto{} were fine-tuned, according to each possible input: no context, tweet, and full context (tweet plus article).
}

\subsection{\hl{Preprocessing}}

\label{sec:preprocessing}

Each tweet was preprocessed using the \emph{pysentimiento} library \cite{perez2021pysentimiento}: we cut character repetitions up to three occurrences; laughs were normalized; user handles were replaced by a special \texttt{@user} token; emojis were converted to a text representation. Hashtags were stripped, surrounded by a special \texttt{hashtag} token, and segmented to words if they were camel-cased.

In order to deal with friendlier computational costs, we limited the sequence lengths to 128, 256, and 512 tokens for the \textbf{None}, \textbf{Tweet} and \textbf{Full} model inputs, respectively.

\hl{
    \subsection{Evaluation}

    We split our dataset into training, development and test sets to train and evaluate our proposed classifiers. To avoid overestimating the performance, we used a disjoint set of articles for the test set. The training and development splits comprise \num{36420} and \num{9120} comments respectively, both coming from 990 articles. The test set has \num{11343} comments from 248 articles.

    Standard metrics were used for both tasks: precision, recall, F1-score and Macro F1 score for the binary classification task. For the fine-grained classification task, we measured F1 for each attacked characteristic, as well as macro-averaged metrics.

}

%% file: src/06_results.tex
\begin{table}[t!]
    \centering
    \normalsize
    \begin{tabular}{l P{0.1\textwidth}  P{0.1\textwidth} P{0.1\textwidth}}
                  & None           & Tweet                & Full           \\
        \hline
        Precision & $71.8 \pm 1.6$ & $\mbf{74.8 \pm 1.9}$ & $72.8 \pm 2.4$ \\
        Recall    & $60.2 \pm 1.4$ & $\mbf{65.3 \pm 1.4}$ & $64.1 \pm 2.3$ \\
        F1        & $65.5 \pm 0.4$ & $\mbf{69.7 \pm 0.3}$ & $68.1 \pm 0.6$ \\
        Macro F1  & $79.8 \pm 0.2$ & $\mbf{82.2 \pm 0.2}$ & $81.3 \pm 0.3$ \\
        \hline
    \end{tabular}

    \caption{Results of classification experiments for the \textbf{binary} detection task. Each model is a \beto{} with three possible inputs: the comment alone without context (\textbf{None}), the comment and the news outlet's tweet (\textbf{Tweet}), and the comment plus the news outlet's tweet plus the article body (\textbf{Full}). Results are expressed as the mean of ten runs of the experiment along with its standard deviation.}
    \label{tab:task_a_results}
\end{table}

Table \ref{tab:task_a_results} displays the results of the binary classification task, measured in accuracy, precision, recall, F1, and Macro F1. Results are expressed as the mean of each metric, along with its standard deviation, over ten independent runs of the experiments. We present the results only for the domain-adapted BETO classifier; full results can be found in Appendix \ref{app:classification_experiments}. We can observe that the model consuming the simple context (\textbf{Tweet}) obtains the best results, with an improvement against the context-unaware (\textbf{None}) model of $4.2$ F1 points on average. The model with the complete context gets worse results than the model with the simple context, although it improves the general performance against the context-unaware version.

\begin{table*}[t]
    \centering
    \normalsize
    \begin{tabular}{l P{0.18\textwidth}  P{0.18\textwidth} P{0.14\textwidth}}
                        & \mc{3}{Context}                                               \\
                        & None            & Tweet                & Full                 \\
        \hline

        CALLS           & $65.1 \pm 1.9$  & $\mbf{68.5 \pm 0.9}$ & $68.0 \pm 1.5$       \\
        POLITICS        & $61.1 \pm 0.8$  & $62.5 \pm 1.3$       & $\mbf{64.8 \pm 1.4}$ \\
        APPEARANCE      & $74.2 \pm 1.0$  & $\mbf{76.6 \pm 0.9}$ & $75.8 \pm 0.9$       \\
        DISABLED        & $58.2 \pm 1.3$  & $\mbf{60.9 \pm 1.8}$ & $57.8 \pm 1.7$       \\
        WOMEN           & $38.9 \pm 1.5$  & $\mbf{42.1 \pm 1.7}$ & $\mbf{42.1 \pm 2.2}$ \\
        RACISM          & $65.3 \pm 1.0$  & $\mbf{72.0 \pm 0.4}$ & $71.1 \pm 1.0$       \\
        CLASS           & $43.3 \pm 1.3$  & $\mbf{51.1 \pm 2.0}$ & $47.6 \pm 2.7$       \\
        LGBTI           & $36.6 \pm 1.9$  & $\mbf{48.2 \pm 1.9}$ & $44.5 \pm 2.1$       \\
        CRIMINAL        & $52.9 \pm 1.1$  & $\mbf{69.9 \pm 1.9}$ & $66.8 \pm 1.7$       \\
        \hline
        Macro F1        & $55.1 \pm 0.5$  & $\mbf{61.3 \pm 0.7}$ & $59.8 \pm 0.6$       \\
        Macro Precision & $63.0 \pm 1.8$  & $\mbf{70.2 \pm 0.9}$ & $67.8 \pm 1.4$       \\
        Macro Recall    & $49.9 \pm 1.2$  & $\mbf{55.1 \pm 1.1}$ & $54.1 \pm 1.3$       \\

        \bottomrule
    \end{tabular}
    \caption{Results of classification experiments for the \emph{fine-grained} task, \hl{measured as F1 score for each of the characteristics and macro-averaged metrics}. Each model is a \beto{} with 3 possible inputs: the analyzed comment alone (\textbf{None}), the comment plus the tweet from the news outlet (\textbf{Tweet}), and comment plus the news outlet's tweet plus the article body (\textbf{Full}). \hl{Results are expressed as the mean of ten runs of the experiment along with its standard deviation.}}
    \label{tab:task_b_results}
\end{table*}

Table \ref{tab:task_b_results} shows the results of the classification experiments for the \textbf{fine-grained} task, measured by F1 score for each of the features and macro-averaged metrics. As expected, the performance boost of including context is more evident in this task, with a difference of approximately 6 points between the context-unaware and context-aware models ($55.1$ vs.\ $61.3$ Macro F1). Regarding the two types of context, again the simple version obtains better performance in practically all the characteristics, with the only exception of POLITICS.

The characteristics that benefit the most from adding context are CRIMINAL ($+17$ F1 points), LGBTI ($+12$), CLASS ($+8$), and RACISM (almost $+7$); on the other hand, APPEARANCE and POLITICS benefit the least. It is worth noting that, even with the help of added context, some characteristics are very difficult for our classifiers and show a relatively low performance: WOMEN, LGBTI and CLASS.

\subsection{Error Analysis}

To have a better understanding of the benefits of adding context and also its limitations, we performed an error analysis between the context-unaware and context-aware models. To do this, we manually checked the output of ten classifiers and looked for their most common errors. Table \ref{tab:error_analysis} shows a selection of test instances where context helps to correctly classify comments, and also some examples where both versions are failing to flag them as hateful. We can observe that context helps to disambiguate some of the messages, which are not clearly understood without the additional information.

A remarkable case is that of LGBTI. The mention of any topic-related word in the headline (such as transgender, gay or lesbian) gives some hint to the classifiers about the nature of the message. Nonetheless, due to the complexity of the offenses to transgender individuals (addressing them by their opposite gender, or slurs about their genitals, for instance) models usually fail in flagging these messages as hateful.

\begin{table*}[ht!]
    \centering
    \begin{tabular}{l l p{0.35\textwidth} p{0.35\textwidth}}
        \toprule
                        &       & Context                                                                                                                                                             & Comment                                                                                                                                                               \\
        \hline

        \multirow{5}{0.10\textwidth}{FN without context, TP  with context}
                        & WOMEN & Ofelia Fernández supported the Government in the controversy over the prisoners and pointed to the Justice that ``hates women''                                     & motherfuck*r ,, hopefully you will soon receive a visit from one of those worms. They will fit you. Willing to support him. Government? Fat creeping larva. Debrained \\
        \rule{0pt}{3ex} & WOMEN & Did More Rial find love in a personal trainer?                                                                                                                      & You have to be hungry to eat that bolivian piglet                                                                                                                     \\
        \rule{0pt}{3ex} & LGBTI & What Elliot Page looks like after declaring himself transgender                                                                                                     & hope she gets psychiatric help                                                                                                                                        \\
        \rule{0pt}{3ex} & LGBTI & Mara Gómez fulfills her dream: she will be the first transgender footballer in the Argentine professional tournament                                                & Mara ``the club'' Gómez                                                                                                                                               \\
        \rule{0pt}{3ex} & LGBTI & Mara Gómez fulfills her dream: she will be the first transgender footballer in the Argentine professional tournament                                                & go break some legs boy                                                                                                                                                \\
        \hline
        \multirow{5}{0.10\textwidth}{FP without context, TN with context}
        \rule{0pt}{3ex} & LGBTI & A man got into his car at the door of the Chinese Embassy and claimed that he had explosives                                                                        & He is not a man. He's a jerk                                                                                                                                          \\
        \rule{0pt}{3ex} & LGBTI & Coronavirus in Argentina: 70\% of cases are in men                                                                                                                  & The corona is female                                                                                                                                                  \\
        \rule{0pt}{3ex} & LGBTI & The ruling party calls for a "federal caravan" in support of the Government and the tax on large fortunes                                                           & Gross                                                                                                                                                                 \\
        \rule{0pt}{3ex} & CLASS & Paul McCartney: "The Chinese need to be cleaner and less medieval"                                                                                                  & it had to be said at last                                                                                                                                             \\
        \rule{0pt}{3ex} & CLASS & Main teaching union rejected the return to presential classes                                                                                                       & shitty bums!                                                                                                                                                          \\
        \hline
        \multirow{6}{0.10\textwidth}{Not detected by any classifier}
        \rule{0pt}{3ex} & WOMEN & Why Women-Led Countries Appear To Have Responded Better To The Coronavirus Crisis                                                                                   & because they wash, iron and sweep?                                                                                                                                    \\
        \rule{0pt}{3ex} & WOMEN & British girl went to Peru for 10 days and stayed for love: she lives with no water and among insects                                                                & she left everything coz of the wood of that Peruvian ahaha that nigger must have a generous dick                                                                      \\
        \rule{0pt}{3ex} & WOMEN & Did More Rial find love in a personal trainer?                                                                                                                      & gotta be well trained to lift that hyppo                                                                                                                              \\
        \rule{0pt}{3ex} & CLASS & The Government will spend \$75B to develop 300 slums throughout the country                                                                                         & without education behind this is nothing, they will remain the same old misfits but now with Netflix.                                                                 \\
        \rule{0pt}{3ex} & LGBTI & She told that she was a lesbian, her father confessed that he was gay and now his mother fell in love with a woman: this is how he was inspired for his second film & The film is called the failure of a normal family                                                                                                                     \\
        \rule{0pt}{3ex} & LGBTI & ``Why don't we see trans doctors?'': The claim of a prestigious cardiologist for America to be more inclusive                                                       & because sick people cannot heal sick people                                                                                                                           \\
        \rule{0pt}{3ex} & LGBTI & A trans woman is killed in Rosario after a burst of 20 shots                                                                                                        & Why did she not pull out her shotgun and apply self-defense?!                                                                                                         \\
        \hline
    \end{tabular}
    \caption{Error analysis between non-contextualized and contextualized classifiers. \hl{Context and comments are shown. }The first group of rows (\emph{FN \hl{---false negatives---} without context, TP \hl{---true positives---} with context}) represent tweets that were incorrectly labeled as non-hateful by non-contextualized classifiers, but contextualized classifiers correctly marked as hateful. The second group consists of tweets that were incorrectly labeled as hateful by non-contextualized classifiers, but contextualized classifiers correctly marked them as non-hateful \hl{(FP stands for false positives and TN for true negatives)}. The last group contains messages that are hateful but were not detected by any classifier, neither non-contextualized nor contextualized. }
    \label{tab:error_analysis}
\end{table*}

%% file: src/07_discussion.tex
\label{sec:discussion}

For the proposed tasks, we can observe that context seems to give a moderate improvement in the binary setting, and a more considerable gain in the fine-grained setting. This result might appear to contradict recent work that found no improvement by means of contextualization in toxicity detection \cite{pavlopoulos2020toxicity}. However, it must be noted that hate speech is one of the most complex forms of toxic behavior; thus, hate speech detection might benefit differently from having additional information.
Also, while \citet{pavlopoulos2020toxicity}'s context was extracted from the entire conversation preceding the target message, our context was taken from the news outlet's tweet and the article itself under discussion.
Further, \citet{xenos-2021-context} recently found that toxicity detection algorithms can take advantage of this additional information by restricting the analysis to a subset of context-sensitive comments.

Something interesting this dataset provides is a characterization of hate speech. Since we have the attacked characteristics for each hateful tweet, we could assess the influence of context for each protected characteristic. Contextual information seems to have more impact on some characteristics than others (e.g., when the attack is against LGBTI people). Moreover, we can observe that the dataset has complex and compositional examples of discriminatory language for some specific characteristics.

The constructed dataset has both short and long contexts. In our experiments, we have observed no substantial improvement in model performance by using the long context; that is, the full article. This might coincide with a familiar behavior observed in humans ---that many people comment after reading nothing but the headline. (However, it might be argued that humans have access to a richer context and information beyond the headline.)

The experiments performed in this work have a few limitations. First, human annotators had access to the full contexts when doing their task. To better assess the impact of context in hate speech detection, context-unaware models should be trained on comments labeled by humans without access to any additional information. Second, a practical limitation is that context is not always available for any given text. Even if we were able to find one, it might not always consist of a news article --- it may also be a conversational thread, or even audiovisual content, for example. Lastly, the labeled comments are replies to tweets published by media outlets, which limits the possible forms of our instances. \hl{Therefore, further study is needed to understand how other forms of messages and contexts impact the detection of hate speech.}

%% file: src/08_conclusions.tex
\label{sec:conclusions}

In this work, we have assessed the impact of adding context to the automatic detection of hate speech. To do this, we built a dataset consisting of user replies to Twitter posts published by main news outlets in Argentina\hl{, and annotated it using carefully designed guidelines. We conducted a series of classification experiments using transformer-based techniques, and found clear evidence that certain contextual information leads to an improved performance: our models showed a 4 to 5 point increase in Macro F1 after adding context}.

Although in our experiments the smallest context (the news article tweet) was the one that obtained the best results, a future line of work could explore ways to include other sources of information. \hl{For instance, adding real-world knowledge about the targets of hate speech could be useful. This information might be even available in the news article itself, or other sources such as a knowledge graph}.

From the error analysis, it can be seen that some categories of hate speech are elusive for state-of-the-art detection algorithms. One of these cases is the abusive messages against the LGBTI community, which contain semantically complex messages, with ironic content and metaphors that are difficult to interpret for classifiers based on state-of-the-art language models. Despite these limitations, the detection of hate speech against the LGBTI community was among the most benefited by the addition of context. \hl{Future work should explore the reasons behind the difficulties for the state-of-the-art models to detect it, and also explore ways to improve the detection of this type of hate speech.}

\hl{We may conclude that hate speech detection clearly benefits from the use of \textbf{contextual information}.} The evidence from our experiments ---preliminary for now, and with the limitations noted in the discussion--- indicates that state-of-the-art models can use this information to improve the detection of hate speech on social networks.
\hl{We hope that this work will encourage the use of contextual information in the detection of hate speech and other opinion mining tasks, and that it will be a starting point for future research in this area.}

%% file: src/biographies.tex
\begin{IEEEbiography}[{\includegraphics[width=1in,height=1.25in,clip, keepaspectratio]{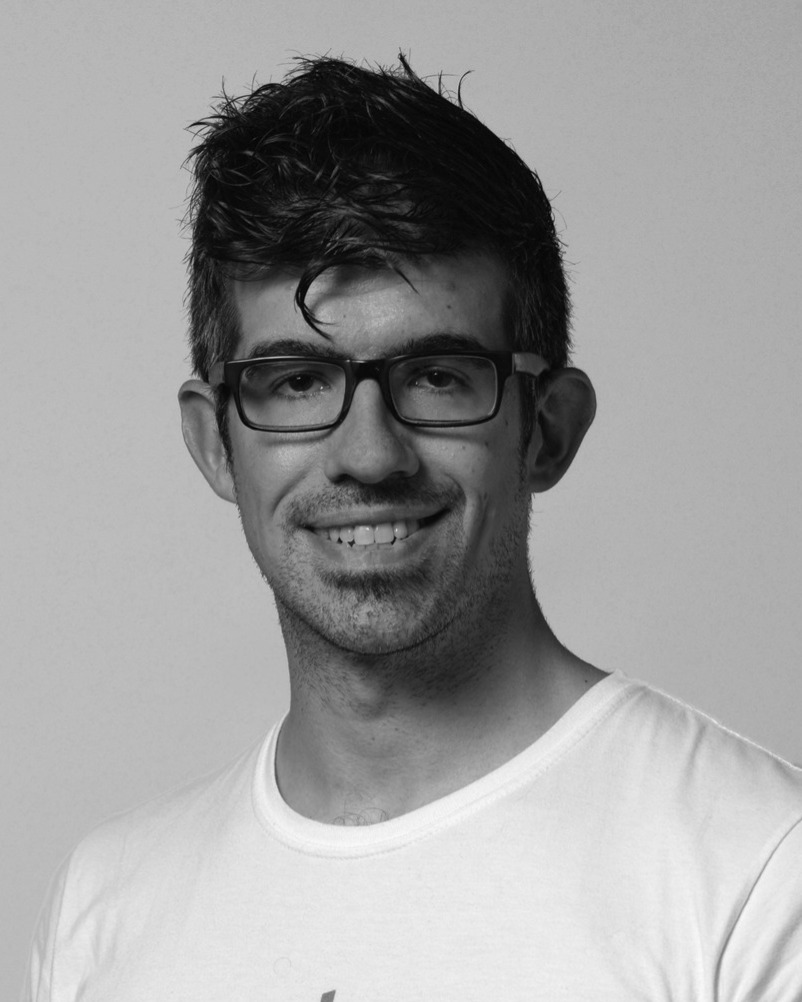}}]{Juan Manuel Pérez} received his MSc (2016) and Ph.D. degree (2022) in computer science at Universidad de Buenos Aires. He is currently a postdoctoral researcher at the Instituto de Ciencias de la Computación, Facultad de Ciencias Exactas, UBA/CONICET, working on contextualized detection of hate speech and opinion mining tasks in social networks using \hl{NLP} techniques.
\end{IEEEbiography}

\begin{IEEEbiography}[{\includegraphics[width=1in,height=1.25in,clip, keepaspectratio]{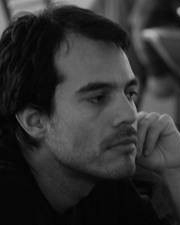}}]{Franco M. Luque} has a degree and a Ph.D. in computer science (FAMAF, National University of Córdoba, Argentina). He works as Adjunct Professor at the Faculty of Mathematics, Astronomy, Physics and Computing (FAMAF), National University of Córdoba, and as an Assistant Researcher at the CONICET. His main area of research is \hl{NLP}, and he is also involved in research projects on various topics such as analysis of hate speech in social networks, multimodal learning for visual dialogue and information extraction in medical reports.
\end{IEEEbiography}

\begin{IEEEbiography}[{\includegraphics[width=1in,height=1.25in,clip, keepaspectratio]{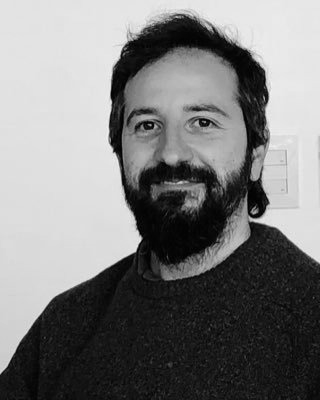}}]{Demian Zayat} is a Constitutional Law Professor. He holds a law degree (2000) from Universidad de Buenos Aires and a master's degree from Stanford University (JSM, 2009). His main area of research is Human Rights and discrimination, combining legal and data-driven research.
\end{IEEEbiography}

\begin{IEEEbiography}[{\includegraphics[width=1in,height=1.25in,clip, keepaspectratio]{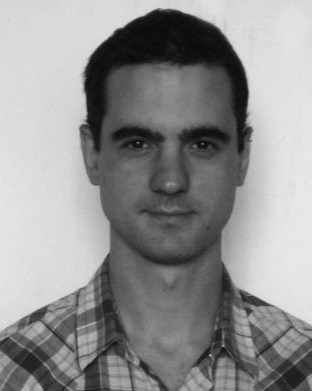}}]{Martín Kondratzky} has a degree in linguistics (Universidad de Buenos Aires). He is currently working as an NLP engineer in the fields of information retrieval and question answering while finishing his master's degree in statistics.
\end{IEEEbiography}

\begin{IEEEbiography}[{\includegraphics[width=1in,height=1.25in,clip, keepaspectratio]{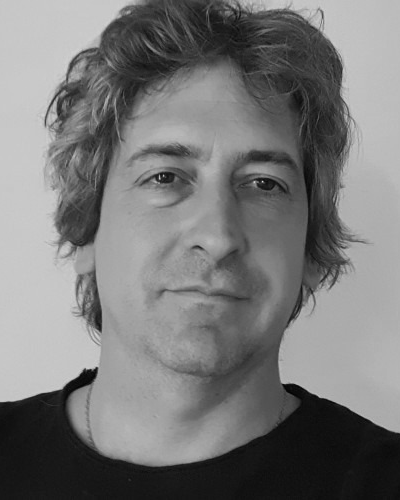}}]{Agustín Moro} holds a Ph.D. in Sociology and is a Professor of Scientific Research Methodology at Universidad Nacional del Centro. He also works as a Consultant in the field of evidence-based policy implementation processes.
\end{IEEEbiography}

\begin{IEEEbiography}[{\includegraphics[width=1in,height=1.25in,clip, keepaspectratio]{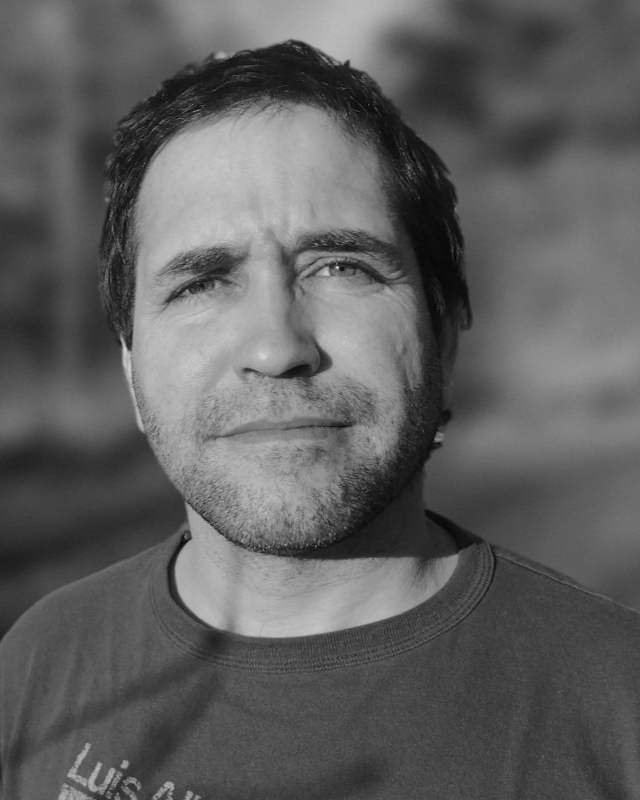}}]{Pablo Santiago Serrati} is a Ph.D. candidate in social sciences at the Faculty of Social Sciences (UBA). He works on the use of data analysis techniques and statistical programming in the fields of Urban Studies and Social Science.
\end{IEEEbiography}

\begin{IEEEbiography}[{\includegraphics[width=1in,height=1.25in,clip, keepaspectratio]{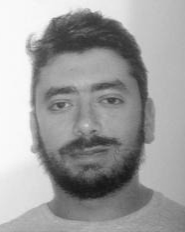}}]{Joaquín Zajac} holds a degree in sociology from Universidad de Buenos Aires, a master's in Social Anthropology and a Ph.D. in social sciences. Post-doc fellow (CONICET) at Universidad de San Martin. He specializes in violence, human rights and security issues, combining qualitative and quantitative data analysis for his research work.
\end{IEEEbiography}

\begin{IEEEbiography}[{\includegraphics[width=1in,height=1.25in,clip, keepaspectratio]{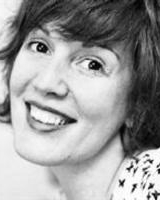}}]{Paula Miguel} holds a degree and a Ph.D. in social sciences. She is a Professor of Sociology and Researcher at the Universidad de Buenos Aires, specializing in cultural analysis, qualitative analysis and, more recently, data science and NLP techniques.
\end{IEEEbiography}

\begin{IEEEbiography}[{\includegraphics[width=1in,height=1.25in,clip, keepaspectratio]{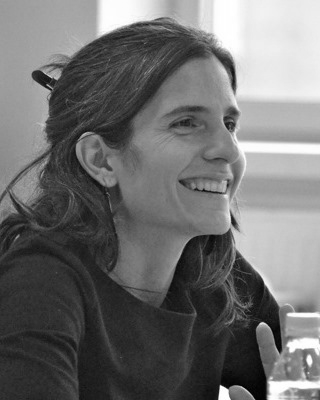}}]{Natalia Debandi} holds a degree in computer science (UBA) and a Ph.D. in Social Sciences from the University of Buenos Aires and the University Paris IV Sorbonne. She specializes in the design of human rights indicators and data analysis for academic and applied human rights research.
\end{IEEEbiography}

\begin{IEEEbiography}[{\includegraphics[width=1in,height=1.25in,clip, keepaspectratio]{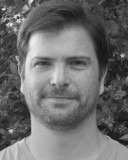}}]{Agustín Gravano} is Associate Professor at Universidad Torcuato Di Tella and Independent Researcher at CONICET, in Argentina. His research focuses on building computational models of coordination in spoken dialogue, for later improving the naturalness of spoken dialogue systems. He received his Ph.D. in computer science from Columbia University in 2009.
\end{IEEEbiography}

\begin{IEEEbiography}[{\includegraphics[width=1in,height=1.25in,clip, keepaspectratio]{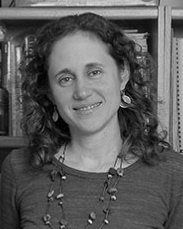}}]{Viviana Cotik} is an Assistant Professor at Universidad de Buenos Aires and a Research Assistant at CONICET, in Argentina. Her research area is NLP and Data Science and focuses on information extraction from biomedical texts. She received her Ph.D. in Computer Science in 2018.
\end{IEEEbiography}

%% file: src/appendix.tex
\subsection{Data selection}
\label{app:data-selection}

\begin{table}[t!]
    \centering
    \begin{tabular}{l l}
        Expression               & Description or translation                      \\
        \hline
        viejo puto               & old fag                                         \\
        marica                   & fag                                             \\
        sodomita                 & sodomite                                        \\
        degenerados              & degenerate                                      \\
        trabuco, trava           & slur for transgender woman                      \\
        travesti                 & transgender woman                               \\
        bija                     & misspelling of dick                             \\
        feministas               & feminists                                       \\
        feminazis                & offensive term against feminists                \\
        aborteras                & abortion activists                              \\
        gorda                    & fat woman                                       \\
        uno menos                & one less (celebratory expression for a killing) \\
        urraca                   & magpie (offensive slur against a woman)         \\
        prostituta               & prostitute                                      \\
        putita                   & little bitch                                    \\
        reventada                & prostitute                                      \\
        peruano, peruca          & peruvian                                        \\
        paraguayo                & paraguayan                                      \\
        trolo                    & fag                                             \\
        bala                     & bullet (as in ``shoot them''); also fag         \\
        bolita                   & slur for bolivian                               \\
        negro(s) (de)            & nigger                                          \\
        judío, sionista          & jew, zionist                                    \\
        matarlos                 & (have to) kill them                             \\
        chinos                   & chinese                                         \\
        una bomba                & a bomb                                          \\
        vayan a laburar/trabajar & go to work                                      \\
        villeros                 & shanty dwellers                                 \\
        \hline
    \end{tabular}
    \caption{Seed expressions used to select articles based on possibly hateful comments.}
    \label{tab:seed_words}
\end{table}

\bigskip

Table \ref{tab:seed_words} lists the seed expressions used to mark potentially hateful comments. This list was constructed manually, checking for some common expressions in the data. We used MongoDB's text index to retrieve any comments containing at least one of them.

Some of these expressions were used literally (with quotation marks) and some were allowed inflections provided by the search engine. For some of them, we excluded other words: for instance, when querying ``negra'' (\emph{female nigger}) we removed ``plata | guita'' (\emph{money}) as there were many hits for such queries. For others, we added prepositions to the query (such as ``negro de'') because using just ``negro'' had a lot of non-hateful hits.

It is important to stress that this method was only used for selecting news articles for the subsequent annotation step, and comments were randomly sampled among the replies to the selected articles.

\subsection{Additional information of the dataset}
\label{app:additional_information}
\begin{table}[t]
    \centering
    \begin{tabular}{lrr}
        Newspaper  & \#Art      & \#Comm      \\
        \hline
        @infobae   & 590        & \num{26834} \\
        @clarincom & 370        & \num{17501} \\
        @LANACION  & 222        & \num{10378} \\
        @cronica   & 42         & \num{ 1562} \\
        @perfilcom & 14         & \num{  594} \\
        \hline
        Total      & \num{1238} & \num{56869} \\
        \hline
    \end{tabular}
    \caption{Number of articles and comments in the dataset per news outlet}
    \label{tab:dataset_per_newspaper}
\end{table}

Table \ref{tab:dataset_per_newspaper} displays the number of articles and comments in the final dataset. We can observe that most articles and comments come from @infobae, followed by @clarincom and @LANACION.

\begin{figure*}[t]
    \centering
    \begin{subfigure}[]{0.45\textwidth}
        \includegraphics[width=\textwidth]{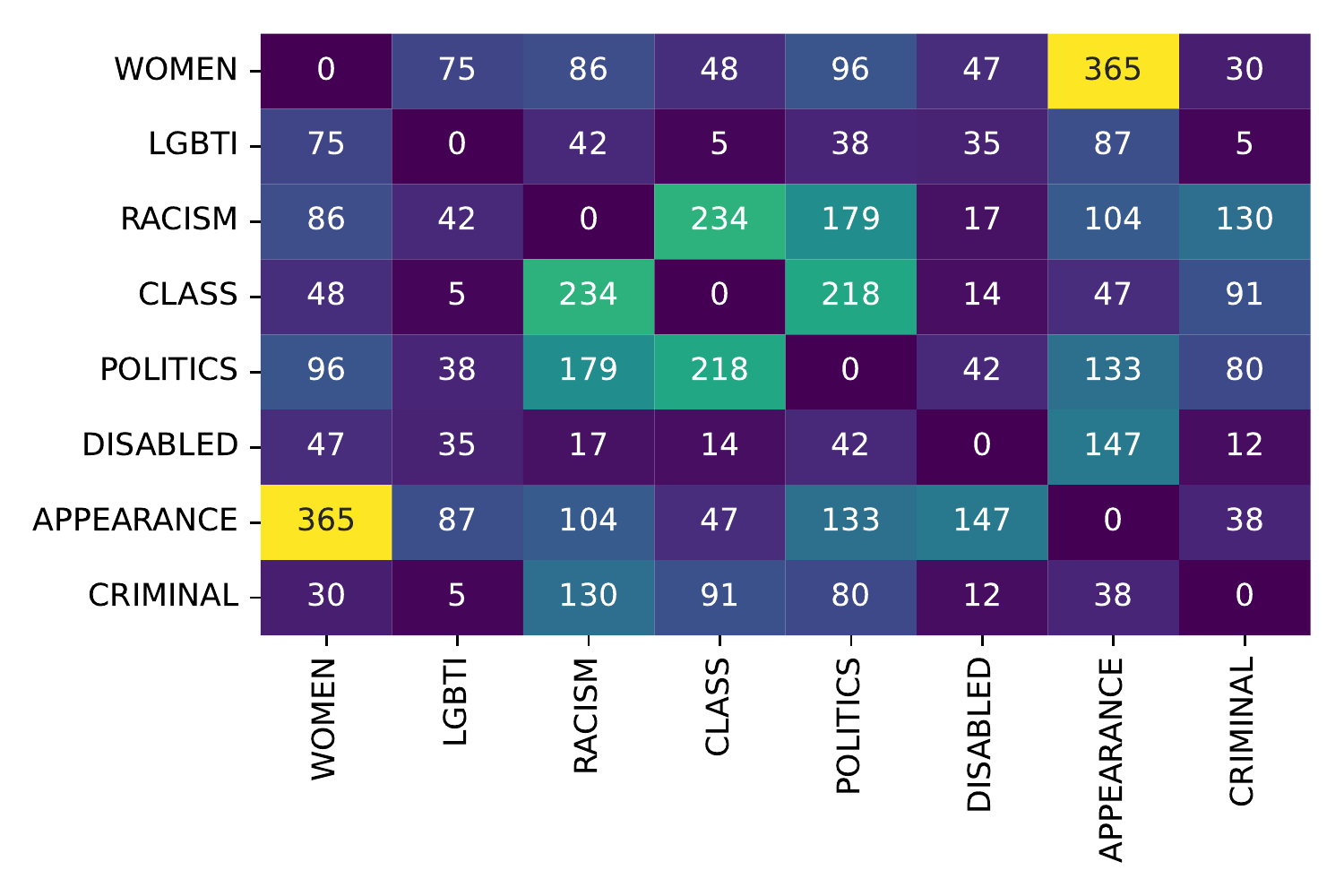}
        \caption{Co-occurrence of attacked characteristics within the same comment}
        \label{subfig:heatmap_characteristics_comment}
    \end{subfigure}
    \begin{subfigure}[]{0.45\textwidth}
        \centering
        \includegraphics[width=\textwidth]{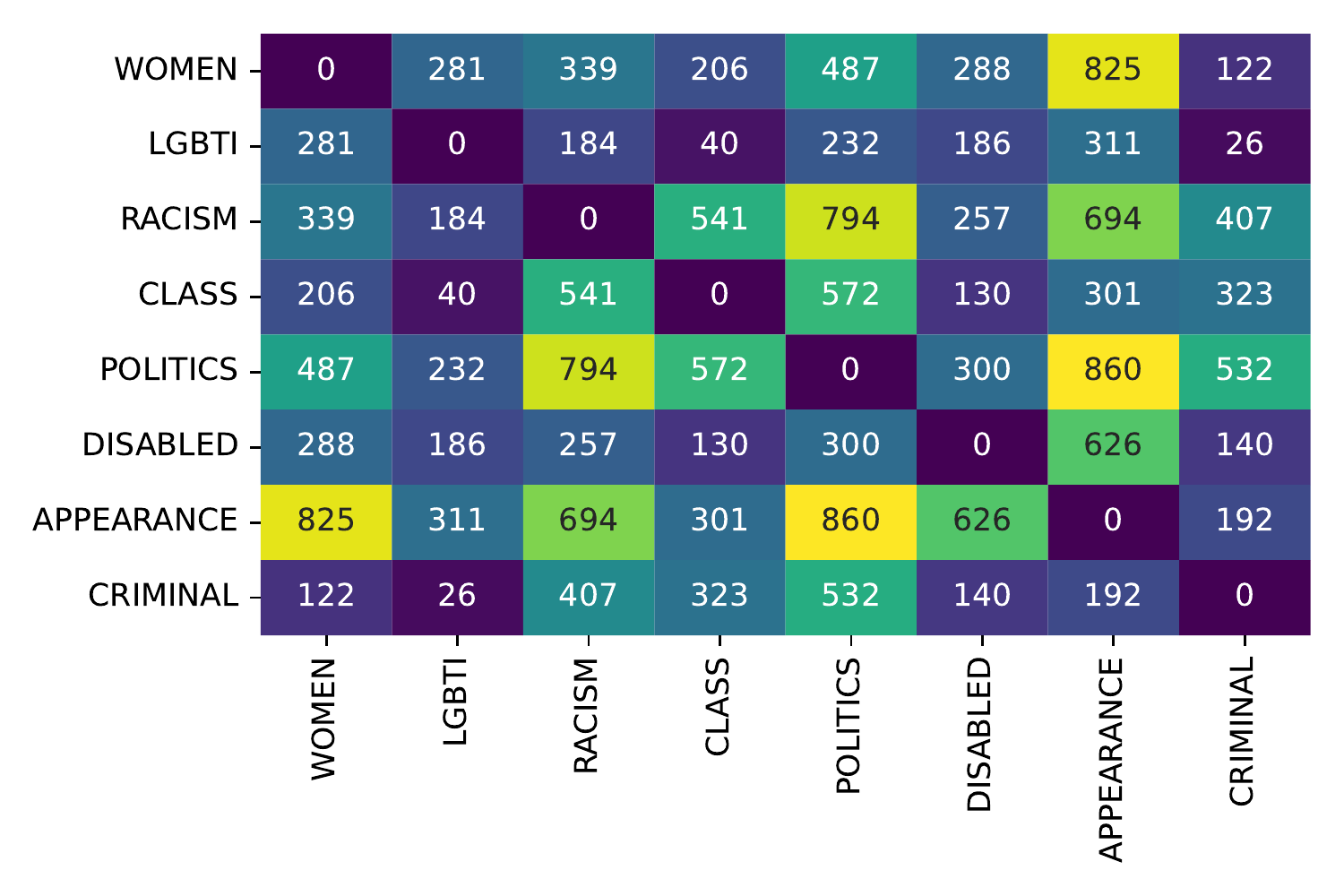}
        \caption{Co-occurrence of attacked characteristics within the same article}
        \label{subfig:heatmap_characteristics_article}
    \end{subfigure}

    \caption{Co-occurrence matrices for attacked characteristics in hateful messages. Figure \ref{subfig:heatmap_characteristics_comment} shows co-occurrence within the same comment, and Figure \ref{subfig:heatmap_characteristics_article} shows co-occurrence across comments of the same article. Brighter indicates more co-occurrence}
    \label{fig:heatmap_characteristics}
\end{figure*}

From the \num{8715} hateful comments present in the dataset, 77\% of them (\num{6777}) contain only one attacked characteristic, nearly 20\% have exactly two, and 220 comments have three or more. Figure \ref{fig:heatmap_characteristics} illustrates the co-occurrence matrix between the different characteristics for comments having more than one attacked characteristic. We can observe that the maximum co-occurrence occurs between the characteristics WOMEN and APPEARANCE, followed by RACISM and CLASS, POLITICS and CLASS, and RACISM and POLITICS.

Another way of analyzing co-occurrence is by grouping the different characteristics of their comments by articles, to observe how the same context can invoke different types of discrimination. Figure \ref{subfig:heatmap_characteristics_article} illustrates the interactions between the different characteristics per article. Greater dispersion is observed in the co-occurrences than in Figure \ref{subfig:heatmap_characteristics_comment}, showing some additional interactions such as between RACISM and POLITICS and ---perhaps unexpectedly--- between APPEARANCE and POLITICS.

\subsection{Classification experiments}
\label{app:classification_experiments}
\begin{table*}
    \centering
    \footnotesize
    \begin{tabular}{l P{0.1\textwidth}P{0.1\textwidth} P{0.1\textwidth}P{0.1\textwidth}  P{0.1\textwidth}P{0.1\textwidth}}
                  & \mc{6}{Context}                                                                                      \\
                  & \mc{2}{None}    & \mc{2}{Tweet}  & \mc{2}{Full}                                                      \\
        Metric    & BETO            & BETO$_{FT}$    & BETO           & BETO$_{FT}$    & BETO           & BETO$_{FT}$    \\
        \hline
        Accuracy  & $88.9 \pm 0.3$  & $89.9 \pm 0.2$ & $90.2 \pm 0.2$ & $91.0 \pm 0.2$ & $90.4 \pm 0.2$ & $90.5 \pm 0.3$ \\
        Precision & $67.8 \pm 2.0$  & $71.8 \pm 1.6$ & $73.1 \pm 1.1$ & $74.8 \pm 1.9$ & $73.9 \pm 1.6$ & $72.8 \pm 2.4$ \\
        Recall    & $56.8 \pm 1.7$  & $60.2 \pm 1.4$ & $60.1 \pm 1.0$ & $65.3 \pm 1.4$ & $61.1 \pm 1.6$ & $64.1 \pm 2.3$ \\
        F1        & $61.8 \pm 0.5$  & $65.5 \pm 0.4$ & $66.0 \pm 0.6$ & $69.7 \pm 0.3$ & $66.9 \pm 0.5$ & $68.1 \pm 0.6$ \\
        Macro F1  & $77.6 \pm 0.3$  & $79.8 \pm 0.2$ & $80.1 \pm 0.3$ & $82.2 \pm 0.2$ & $80.6 \pm 0.2$ & $81.3 \pm 0.3$ \\
        \bottomrule
    \end{tabular}
    \caption{Results of the classifiers for the binary task, expressed as the mean and standard deviation of ten independent runs of the experiments. Three different types of inputs are considered: no context, comment + tweet of the news outlet, and full context. FT means that the pre-trained language model was fine-tuned, and $\neg$FT means it was not fine-tuned.}
    \label{tab:full_plain_results}
\end{table*}

Table \ref{tab:full_plain_results} and Table \ref{tab:full_results} display the full results for the binary and fine-grained tasks. We used two pre-trained language models as our base models: BETO, without any fine-tuning on the data (marked as $\neg$FT), and a BETO fine-tuned with the remaining data of the collection process, as described in Section \ref{sec:classification_algorithms}. The results show that, in all cases, the fine-tuning process improves the performance of the classifiers.

\hl{
    To train our classification models, we used the \emph{HuggingFace} library \cite{wolf2020transformers} and the \emph{PyTorch} framework \cite{paszke2019pytorch}. We used a \emph{NVIDIA GeFORCE GTX 1080 Ti} to fine-tune the models. To perform the domain-adaptation of the language models, we used a \emph{TPU v2-8} in a \emph{Google Colab Pro} instance, taking 10 hours at its maximum sequence length.
}

\begin{table*}
    \centering
    \footnotesize
    \begin{tabular}{l P{0.1\textwidth}P{0.1\textwidth} P{0.1\textwidth}P{0.1\textwidth}  P{0.1\textwidth}P{0.1\textwidth}}
                        & \mc{6}{Context}                                                                                      \\
                        & \mc{2}{None}    & \mc{2}{Tweet}  & \mc{2}{Full}                                                      \\
        Metric          & $\neg$FT        & FT             & $\neg$FT       & FT             & $\neg$ FT      & FT             \\
        \hline
        CALLS           & $64.6 \pm 1.0$  & $65.1 \pm 1.9$ & $63.8 \pm 0.9$ & $68.5 \pm 0.9$ & $65.3 \pm 1.3$ & $68.0 \pm 1.5$ \\
        WOMEN           & $37.3 \pm 1.3$  & $38.9 \pm 1.5$ & $41.1 \pm 0.9$ & $42.1 \pm 1.7$ & $38.1 \pm 1.7$ & $42.1 \pm 2.2$ \\
        LGBTI           & $35.1 \pm 1.8$  & $36.6 \pm 1.9$ & $45.1 \pm 2.1$ & $48.2 \pm 1.9$ & $42.7 \pm 2.4$ & $44.5 \pm 2.1$ \\
        RACISM          & $63.5 \pm 1.4$  & $65.3 \pm 1.0$ & $68.8 \pm 1.2$ & $72.0 \pm 0.4$ & $69.1 \pm 0.9$ & $71.1 \pm 1.0$ \\
        CLASS           & $40.1 \pm 1.6$  & $43.3 \pm 1.3$ & $49.1 \pm 2.2$ & $51.1 \pm 2.0$ & $45.1 \pm 1.9$ & $47.6 \pm 2.7$ \\
        POLITICS        & $55.5 \pm 1.8$  & $61.1 \pm 0.8$ & $57.9 \pm 1.4$ & $62.5 \pm 1.3$ & $59.1 \pm 1.3$ & $64.8 \pm 1.4$ \\
        DISABLED        & $55.1 \pm 1.6$  & $58.2 \pm 1.3$ & $58.5 \pm 1.6$ & $60.9 \pm 1.8$ & $55.7 \pm 2.3$ & $57.8 \pm 1.7$ \\
        APPEARANCE      & $72.6 \pm 1.0$  & $74.2 \pm 1.0$ & $74.1 \pm 1.2$ & $76.6 \pm 0.9$ & $75.5 \pm 0.9$ & $75.8 \pm 0.9$ \\
        CRIMINAL        & $51.3 \pm 1.4$  & $52.9 \pm 1.1$ & $65.0 \pm 1.2$ & $69.9 \pm 1.9$ & $65.4 \pm 2.3$ & $66.8 \pm 1.7$ \\
        \hline
        Macro Precision & $55.8 \pm 1.0$  & $63.0 \pm 1.8$ & $64.2 \pm 1.6$ & $70.2 \pm 0.9$ & $67.7 \pm 1.4$ & $67.8 \pm 1.4$ \\
        Macro Recall    & $50.6 \pm 0.6$  & $49.9 \pm 1.2$ & $54.0 \pm 0.8$ & $55.1 \pm 1.1$ & $50.4 \pm 0.9$ & $54.1 \pm 1.3$ \\
        Macro F1        & $52.8 \pm 0.5$  & $55.1 \pm 0.5$ & $58.2 \pm 0.5$ & $61.3 \pm 0.7$ & $57.3 \pm 0.7$ & $59.8 \pm 0.6$ \\
        \bottomrule
    \end{tabular}
    \caption{Results of the classifiers for the fine-grained task, expressed as the mean and standard deviation of ten independent runs of the experiments. Three different types of inputs are considered: no context, comment + tweet of the news outlet, and full context. FT means that the pre-trained language model was fine-tuned, and $\neg$FT means it was not fine-tuned. }
    \label{tab:full_results}
\end{table*}